\pdfoutput=1
\documentclass[preprint,11pt]{elsarticle}

\usepackage[utf8]{inputenc}
\usepackage[T1]{fontenc}
\usepackage{times}
\usepackage[margin=1in]{geometry}
\usepackage{setspace}
\usepackage{graphicx}
\usepackage{booktabs}
\usepackage{array}
\usepackage{amsmath,amssymb}
\usepackage{hyperref}
\AtBeginDocument{}
\usepackage{xcolor}
\usepackage{microtype}
\usepackage{float}
\usepackage{tikz}
\usetikzlibrary{shapes.geometric, arrows.meta, positioning, calc, fit, backgrounds, decorations.pathreplacing}
\usepackage{algorithm}
\usepackage{algpseudocode}
\usepackage{pgfplots}
\usepgfplotslibrary{statistics}
\usepackage{subcaption}

\hypersetup{
    colorlinks=true,
    linkcolor=blue!70!black,
    citecolor=blue!70!black,
    urlcolor=blue!70!black,
    pdftitle={A Two-Scan Deep Learning Model for Predicting Dementia in Mild Cognitive Impairment},
    pdfauthor={Sara Fin, Alireza Moayedikia, David J. White, Uffe Kock Wiil, Alicia Troncoso},
}

\usepackage{pgfplotstable}
\usepackage{multirow}
\pgfplotsset{compat=1.17}
\usepackage{bbm}
\usetikzlibrary{decorations.pathreplacing}

\makeatletter
\def\ps@pprintTitle{\let\@oddhead\@empty\let\@evenhead\@empty\def\@oddfoot{}\let\@evenfoot\@oddfoot}
\makeatother

\begin{document}
\sloppy

\begin{frontmatter}

\title{A Two-Scan Deep Learning Model for Predicting Dementia in Mild Cognitive Impairment}

\author[1]{Sara Fin}

\author[2]{Alireza Moayedikia\corref{cor1}}
\cortext[cor1]{Corresponding author: Alireza Moayedikia (PhD) is with the Department of Business Technology and Entrepreneurship, School of Business Law and Entrepreneurship, Swinburne University of Technology, John Street, Hawthorn 3122, VIC, Australia -- Email: amoayedikia@swin.edu.au}

\author[3]{David J. White}

\author[4]{Uffe Kock Wiil}

\author[5]{Alicia Troncoso}

\makeatletter
\g@addto@macro\elsauthors{\def\authorsep{\\[3pt]\normalfont}}
\makeatother
\author[]{for the Alzheimer's Disease Neuroimaging Initiative\fnref{adni}}
\fntext[adni]{Data used in preparation of this article were obtained from the Alzheimer's Disease Neuroimaging Initiative (ADNI) database (adni.loni.usc.edu). As such, the investigators within the ADNI contributed to the design and implementation of ADNI and/or provided data but did not participate in analysis or writing of this report. A complete listing of ADNI investigators can be found at: \url{http://adni.loni.usc.edu/wp-content/uploads/how_to_apply/ADNI_Acknowledgement_List.pdf}}

\affiliation[1]{
    organization={Australian Regenerative Medicine Institute, Monash University},
    city={Melbourne},
    state={VIC},
    country={Australia}
}

\affiliation[2]{
    organization={School of Business Law and Entrepreneurship, Swinburne University of Technology},
    city={Melbourne},
    state={VIC},
    country={Australia}
}

\affiliation[3]{
    organization={Swinburne Neuroimaging, Centre for Mental Health and Brain Sciences, School of Health Sciences, Swinburne University of Technology},
    city={Melbourne},
    state={VIC},
    country={Australia}
}

\affiliation[4]{
    organization={The Maersk Mc-Kinney M{\o}ller Institute, University of Southern Denmark},
    city={Odense},
    country={Denmark}
}

\affiliation[5]{
    organization={Data Science \& Big Data Lab, Universidad Pablo de Olavide},
    city={Seville},
    country={Spain}
}

\begin{abstract}
Predicting which people with mild cognitive impairment (MCI) will develop dementia matters for starting treatment early, yet computational work on structural brain imaging has relied almost entirely on a single scan. We propose TAFNet, a temporal attention fusion network that combines a baseline and a follow-up T1-weighted scan. A pretrained Siamese encoder represents each scan, and a fusion module combines the two through anatomical difference, cross-temporal attention and joint context, mixed by a learned per-patient gate. We evaluate it on paired scans from participants with MCI in the Alzheimer's Disease Neuroimaging Initiative, taken up to two years apart, with a dementia diagnosis within three years of the first scan as the outcome. The evaluation is designed to hold up under scrutiny: conversion is defined from recorded diagnoses, the pretraining pool shares no participants with the evaluation cohort, a held-out test partition is kept apart from model development, and uncertainty is estimated by resampling participants. TAFNet discriminates converters from non-converters well on the held-out partition. It outperforms conventional single-scan networks and a model that subtracts the two scans; both differences are significant in cross-validation and have the same direction on the smaller held-out partition. With everything else held fixed, adding the follow-up scan gives a significant gain in cross-validation. Against a recurrent CNN-LSTM baseline built on the same encoder, TAFNet performs comparably, with neither model consistently ahead. Operating points chosen on validation data, in place of the default threshold, give high sensitivity at a clinically reasonable specificity.
\end{abstract}


\end{frontmatter}

\newpage
\section{Introduction}

Alzheimer's disease (AD) is a progressive brain disorder that gradually destroys memory and thinking, and it is the most common cause of dementia worldwide~\cite{scheltens2021alzheimer}. The disease begins silently, often years or even decades before symptoms appear. During this hidden phase two abnormal proteins build up in the brain, amyloid-$\beta$ plaques and tau tangles, and this build-up eventually causes widespread loss of neurons~\cite{jack2010hypothetical}. Amyloid appears before tau, but it is tau and the brain shrinkage that accompanies it that track cognitive decline most closely~\cite{jack2010hypothetical,jack2013tracking}. Many older people with normal cognition already carry substantial amyloid pathology, which shows that the preclinical phase is long. That long phase is both a diagnostic problem and a treatment opportunity~\cite{jack2010hypothetical}.

Because AD develops over such a long period, it is described in stages. In the preclinical stage, people are cognitively normal but already show biological signs of the disease through biomarkers of amyloid, neurodegeneration, or both~\cite{sperling2011preclinical}. The next stage is mild cognitive impairment (MCI). People with MCI notice changes in their thinking and have measurable impairment, usually in memory, but they still manage daily life on their own and do not meet the criteria for dementia~\cite{albert2011mci}. MCI is common in older adults and is the stage at which treatment is most likely to help~\cite{petersen2014mci}.

The clinical problem is that MCI does not have a single outcome. Meta-analyses show that many people with MCI progress to dementia, many remain stable, and some return to normal cognition~\cite{mitchell2009rate}. Reported conversion rates differ between specialist clinics and community settings, and also depend on patient characteristics and on how the diagnosis is made~\cite{ward2013rate}. The amnestic subtype usually progresses to AD, while non-amnestic subtypes may lead to other forms of dementia~\cite{petersen2014mci}. MCI is therefore not simply early dementia, and tools that predict who will convert are needed to identify the people early treatment would help most.

Structural magnetic resonance imaging (MRI) is a useful way to detect and track brain degeneration in AD. Shrinkage follows a predictable route: it begins in the medial temporal lobe, especially the hippocampus and entorhinal cortex, then spreads to the temporal neocortex, posterior cingulate and precuneus, and eventually reaches much of the cortex~\cite{frisoni2010clinical}. This route mirrors the spread of tau pathology, which is why MRI can be used to stage the disease and to predict outcomes.

The hippocampus and entorhinal cortex have drawn the most attention as early markers. Prospective studies show that smaller volumes in these regions predict conversion from MCI to AD on their own, and the hippocampus predicts more strongly than the entorhinal cortex~\cite{devanand2007hippocampal}. These volumes fall in a clear order across groups: largest in cognitively normal controls, smaller in stable MCI, and smallest in people with MCI who later convert~\cite{devanand2007hippocampal}. Measures of cortical thickness show a consistent ``AD signature'' covering the medial temporal cortex, temporal pole, inferior temporal gyrus, lateral parietal cortex and precuneus~\cite{dickerson2009cortical}. This signature appears not only in mild AD dementia but also in cognitively normal older adults with amyloid pathology~\cite{dickerson2009cortical}. The relationship between cortical thickness and early amyloid is not simple, however: recent longitudinal work reports that thickness changes precede high amyloid levels by at least seven years and are not limited to thinning~\cite{roe2026cortical}. Cortical thickness is therefore a candidate early marker of later decline rather than a settled one.

Longitudinal MRI studies show that the \emph{rate} of shrinkage carries information a single scan cannot. In healthy older people, faster ventricular expansion predicts conversion to MCI or AD; in people with MCI, both ventricular and whole-brain shrinkage rates predict later conversion to AD~\cite{jack2005brain}. Baseline hippocampal volume and the rate of shrinkage add to each other, because prognosis depends both on how far degeneration has gone and on how fast it is moving~\cite{jack2005brain}.

Large multicentre efforts, above all the Alzheimer's Disease Neuroimaging Initiative (ADNI), have confirmed these findings. Work on this public dataset shows that ventricular enlargement is an objective and sensitive marker of progression, with AD participants expanding far more over six months than controls~\cite{nestor2008ventricular}. Tracking the same cohorts over time also shows that people with MCI who later convert shrink almost as fast as people with established AD, particularly in the hippocampus and entorhinal cortex~\cite{risacher2010longitudinal}. Reviews now treat longitudinal structural MRI as the most efficient outcome measure for clinical trials, ahead of cognitive tests~\cite{weiner2017recent}.

Several kinds of biomarker have been studied for predicting MCI-to-AD progression, including cerebrospinal fluid markers, blood assays and positron emission tomography~\cite{scheltens2021alzheimer,ossenkoppele2022tau}. Some of these predict well, but they cost more, require invasive collection, or are hard to obtain outside academic centres. Structural MRI is widely available, non-invasive and already part of routine care, which makes it a good basis for prediction tools that could be used at scale. It has, however, been studied mostly one scan at a time, and a single scan cannot capture the fact that AD is a process that unfolds.

Clinical and biomarker research has established the value of structural MRI, but the computational analysis of these images has mostly used a single snapshot. Early machine learning work separated AD from healthy controls well, yet predicting which people with MCI will progress to AD dementia has proved much harder. The gap has two causes: the changes at the prodromal stage are subtle, and the MCI population is biologically varied. Studies using features that transfer easily to the clinic---cognitive scores, hippocampal volumes and APOE status---keep finding the same ``hard to classify'' subgroup, which suggests the variation is real rather than a modelling artefact~\cite{rye2022predicting}.

The main limitation of these studies is that they use baseline imaging alone. Work that addresses this directly shows that longitudinal models, such as convolutional neural network (CNN) embeddings combined with time-aware long short-term memory (LSTM) networks for survival analysis, do substantially better than single-timepoint models~\cite{aghajanian2025longitudinal}. Patterns of change over time therefore carry prognostic information that one scan cannot provide. These early longitudinal studies, though, used older architectures with no way of modelling relationships between distant brain regions, so there is room to improve the architecture.

The evidence above leaves two lines of work unjoined. On the clinical side, rates of shrinkage predict better than single-scan measures \cite{jack2005brain,risacher2010longitudinal}, and computational work confirms that modelling change over time improves prediction \cite{aghajanian2025longitudinal}. On the methodological side, hybrid CNN-Transformer architectures are the current state of the art for AD classification and reach high accuracy under careful evaluation \cite{hu2023convswinformer,zhao2024vecnn,zhou2025deep}.

The two have not been brought together. Longitudinal models still use older architectures that cannot relate distant brain regions, and modern hybrid CNN-Transformer models have only been applied to single scans. This study closes that gap and makes three contributions.

First, we propose TAFNet, a hybrid CNN-Transformer with temporal attention that takes a pair of scans. A pretrained Siamese encoder represents each timepoint, and a Temporal Fusion Module combines the two through three branches, anatomical difference, cross-temporal attention and joint context, mixed by a learned per-patient gate. Second, we set out a cohort and labelling protocol for longitudinal ADNI analysis. Conversion is taken from recorded diagnoses rather than from enrolment category, censored participants are excluded rather than counted as stable, and the pool used to pretrain the encoder is checked to share no participants with the evaluation cohort. Third, we evaluate the method under a single protocol against single-timepoint convolutional baselines, against subtraction of the two scans, and against a different longitudinal aggregator, with participant-clustered inference and a held-out partition. TAFNet outperforms the single-scan and subtraction baselines in cross-validation, with the same direction on the smaller held-out partition, and the follow-up scan adds a measurable gain within the model. Against a recurrent CNN-LSTM aggregator built on the same encoder it is comparable, and we report that comparison in full. We also test two claims the design invites and find neither supported: ablations show that no single fusion branch is necessary, and the gate coefficients do not explain individual predictions.

\section{Literature Review}
\label{sec:literature_review}

Deep learning has changed how Alzheimer's disease is classified and predicted from structural MRI. What an architecture needs in order to diagnose AD from one scan, however, is not what it needs in order to predict decline from changes in shape over time. This section reviews the field in two parts, so that the contribution of the proposed Temporal Attention Fusion Network (TAFNet) can be placed against it. The first part follows single-timepoint models from standard 3D convolutional networks through Vision Transformers to hybrid designs. The second part looks at longitudinal architectures and at the design choices---where the two scans are merged, and by what mechanism---that decide how a pair of images is combined.

Early deep learning work on AD classification had methodological problems, above all data leakage in 2D slice-based designs, which inflated the accuracy they reported~\cite{wen2020cnn}. Later work using subject-level cross-validation and full 3D volumes has shown real improvements. Lightweight 3D CNNs with dilated convolutions separate AD from cognitively normal participants well, and were validated externally on several datasets~\cite{yee2021construction}. Custom 3D CNNs validated on the National Alzheimer's Coordinating Center (NACC) cohort separated cognitively normal participants from those with MCI or mild AD~\cite{liu2022generalizable}. In that study, people with MCI whom the model wrongly labelled as AD went on to progress faster in reality, which suggests the predictions carry prognostic as well as diagnostic information~\cite{liu2022generalizable}. Multi-scale attention-based pseudo-3D CNNs trade some volumetric processing for lower memory use and still perform competitively~\cite{pei2022multiscale}. Whole-brain 3D CNNs that take the complete volume without selecting features first show that end-to-end learning can find useful patterns outside the regions usually examined~\cite{folego2020alzheimer}.

Generalising to new clinical settings remains an unsolved problem. Ensembles of several architectures report high accuracy on ADNI but lose accuracy consistently when validated on outside data~\cite{fathi2024ensemble}. CNNs are also good at local spatial features and poor at relating regions that are far apart, which matters for AD because its atrophy pattern is spread across several structures. Vision Transformers (ViTs) treat an image as a sequence of patches and use self-attention, so they can model those long-range relationships. Pure transformers, however, need large training sets to work well, which is a problem in medical imaging.

Hybrid designs combine the two approaches. Conv-Swinformer puts VGGNet-16 in front of a Swin Transformer encoder and classifies AD accurately, with reasonable accuracy on the harder task of separating cognitively normal participants from MCI~\cite{hu2023convswinformer}. Vision-transformer-equipped CNNs (VECNN) perform well on three-way classification and are notably sensitive to MCI~\cite{zhao2024vecnn}. Residual depthwise separable 3D CNNs combined with Video Swin Transformers classify AD very accurately, and the same study shows clearly how much 2D slice-based evaluation inflates accuracy relative to full volumes~\cite{zhou2025deep}. For MCI-to-AD prediction specifically, Vision Transformers applied to mid-sagittal slices work reasonably well, although a single slice discards most of the spatial information in the volume~\cite{hoang2023vision}.

Self-supervised learning has been used to work around the shortage of labelled scans. Masked autoencoders (MAE) pretrained on non-AD brain MRI and then fine-tuned on ADNI transfer effectively~\cite{kunanbayev2024training}, and contrastive pretraining matches supervised baselines on balanced accuracy~\cite{gryshchuk2025contrastive}. These methods look promising, but they have rarely been applied to longitudinal, multi-timepoint MCI prediction.

Single-timepoint models have moved the field forward, but longitudinal structural MRI carries prognostic information that a baseline scan does not. This subsection looks at how pairs of volumes are actually processed, and at three linked choices: at what stage the two timepoints are merged, whether the two encoding streams share weights, and what operation combines the extracted features.

The simplest choice is early fusion, which stacks the baseline and follow-up volumes as channels before any features are extracted. It is easy to implement, but it forces the first layers to extract spatial features and compare timepoints at the same time, which makes the network sensitive to registration error. Late fusion instead combines the predictions of two independent pathways. This avoids the problem but cannot model any interaction between the two timepoints. The field has therefore settled on intermediate fusion~\cite{martnezmucia2021deep,qiu2023multimodal3dsiamesenet,chen2024longformer}. Ocasio and Duong~\cite{martnezmucia2021deep}, for example, found a twin encoder with a concatenation merge to be the best of the longitudinal configurations they tried. Here, two parallel branches extract spatial representations separately and merge them part-way through the network, so the merging layers work on abstract features rather than on raw voxels.

Within intermediate fusion, a true Siamese network uses two identical branches with shared weights, which guarantees that both volumes land in the same feature space~\cite{qiu2023multimodal3dsiamesenet}. Ostertag et al.~\cite{qiu2023multimodal3dsiamesenet} showed that this isolates change in shape cleanly: any difference computed between the two representations reflects real anatomical change rather than a quirk of one encoder. Sharing weights also halves the parameter count and so acts as a regulariser. Dropping the constraint, in ``twin'' configurations, lets the baseline encoder specialise in fixed structural vulnerability while the follow-up encoder tracks active degeneration. This improves accuracy but raises the risk of overfitting~\cite{martnezmucia2021deep}.

Once the representations exist, the merge operation decides what temporal information the rest of the network can use. Element-wise subtraction isolates how much structure has been lost, which matches the clinical definition of neurodegeneration~\cite{qiu2023multimodal3dsiamesenet}, but it discards the baseline context~\cite{martnezmucia2021deep}. Concatenation keeps that context~\cite{martnezmucia2021deep} but leaves the dense layers downstream to discover the temporal relationship, and those layers have no geometric guidance for aligning two 3D grids. Recent architectures use cross-attention instead. Chen et al.~\cite{chen2024longformer} built LongFormer, which applies cross-attention to the longitudinal pair to handle spatial misalignment. The same mechanism has been used across modalities rather than across time: Zhang et al.~\cite{zhang2025cross} fuse structural MRI with Jacobian determinant maps at one timepoint using asymmetric, bidirectional cross-attention. Conditioning features explicitly on the interval between scans has also been shown to improve predictions of the progression trajectory~\cite{ouyang2025stamp}.

These architectures move steadily from fixed arithmetic towards adaptive, attention-driven fusion, and each mechanism gives up something. Subtraction shows change but loses context, concatenation keeps both timepoints but does not align them, and cross-attention aligns them but gives no explicit measure of how much has changed. No existing architecture provides all three views of the temporal relationship within a single learnable fusion mechanism, which is what the architecture proposed in Section~\ref{sec:proposed_method} is designed to do.

\section{Preprocessing}
\label{sec:preprocessing}

This section describes the data source and the preprocessing pipeline applied to every structural MRI volume before training\footnote{Source code available at \url{https://github.com/amoayedikia/tafnet}.}. Careful preprocessing matters in neuroimaging deep learning for three reasons: the same anatomy must sit in the same place across participants, non-brain tissue must be removed, and intensity ranges, which differ substantially between scanners, must be brought onto a common scale.

\subsection{Imaging Data}
\label{sec:dataset}

Data used in this study were obtained from the Alzheimer's Disease Neuroimaging Initiative (ADNI) database (\url{adni.loni.usc.edu}). The ADNI was launched in 2003 as a public-private partnership, led by Principal Investigator Michael W.\ Weiner, MD. The primary goal of ADNI has been to test whether serial MRI, positron emission tomography, other biological markers, and clinical and neuropsychological assessment can be combined to measure the progression of MCI and early AD. For up-to-date information on the initiative, see \url{www.adni-info.org}.

Two separate sets of participants are drawn from ADNI. The longitudinal cohort consists of participants with an MCI diagnosis at baseline and at least one qualifying pair of T1-weighted scans; every result in this paper comes from it. A second, cross-sectional pool of cognitively normal and dementia participants is used only to pretrain the encoder. Conversion labels come from the ADNI diagnostic summary table rather than from enrolment category, censored participants are excluded rather than counted as stable, and the two sets are checked to share no participants. These choices determine both the size of the cohort and the prevalence of conversion. They are set out in full in Section~\ref{sec:cohort_protocol}, together with the exclusion counts, the resulting partitions and the evaluation protocol. The present section covers only the imaging data and the preprocessing, which are the same for both sets.

\label{sec:acquisition}
The scans were acquired between 2006 and 2017, at 54 sites for the longitudinal cohort and 59 sites for the pretraining pool, under the ADNI MRI protocols in force at the time~\cite{jack2008adni}. The data are therefore not homogeneous in vendor, sequence or acceleration. All 945 scans in the longitudinal cohort were acquired at 3\,T, which we confirmed from the DICOM headers: 561 on Siemens, 230 on GE and 154 on Philips systems. All volumes are 3D T1-weighted, acquired as 1.2\,mm sagittal slices with an in-plane resolution of 1.0 to 1.05\,mm, but the sequence depends on the scanner vendor. Siemens and Philips systems, and 16 scans from older GE systems, used magnetisation-prepared rapid gradient echo (MPRAGE; nominal TR $\approx 2300$\,ms, TE $\approx 3$\,ms, TI $\approx 900$\,ms, flip angle $9^\circ$). The remaining GE scans used inversion-recovery spoiled gradient echo (IR-SPGR or IR-FSPGR). Of the 945 scans in the longitudinal cohort, 731 are MPRAGE and 214 are IR-(F)SPGR, and 412 were acquired with parallel-imaging acceleration; Table~\ref{tab:acquisition} gives the breakdown for both participant sets. The protocol also varies within participants: in 284 of the 604 pairs one scan is accelerated and the other is not, in 13 pairs the two scans come from different sequence families, in 15 from different vendors, and in 41 from different scanner models. Conversion is about as common in the mixed-acceleration pairs (19.4\%) as in pairs with consistent acceleration (24.4\%). The same check on the 1{,}948 scans of the pretraining pool, read from the image headers at conversion, gave the same result: all were acquired at 3\,T, 1{,}051 on Siemens, 552 on GE and 345 on Philips systems. We downloaded the original, unprocessed ADNI images, so no gradient-warp, B1 or N3 correction had been applied upstream. Raw DICOM images were converted to NIfTI format using \texttt{dcm2niix} (version 1.0.20211006)~\cite{li2016dcm2niix}, applying BIDS-compliant naming conventions.

\subsection{Preprocessing Pipeline}
\label{sec:pipeline}

Every T1-weighted volume was processed with a longitudinal pipeline built on Advanced Normalisation Tools (ANTs, through ANTsPy 0.6.3)~\cite{avants2011ants} and ANTsPyNet 0.3.2~\cite{tustison2021antsx}. The pipeline is shown in Figure~\ref{fig:preprocessing} and stated formally in Algorithm~\ref{alg:preprocessing}. It follows the practice of dedicated longitudinal processing streams, in which all scans of a participant are first aligned to an unbiased within-subject template, and only that template is registered to standard space~\cite{reuter2012within}.

Three properties of the design matter for a model that compares two timepoints. First, the two scans of a pair are treated identically, and each is interpolated exactly once. Second, all timepoints of a participant share one brain mask and one transformation to standard space, so a difference between timepoints cannot come from separately estimated masks or separately estimated template registrations. Third, the whole brain lies inside the network input. The same pipeline is applied to the pretraining pool, where a participant with a single scan (91 of the 531 pool participants) serves as their own template.

\begin{figure}[H]
\centering
\resizebox{\linewidth}{!}{%
\begin{tikzpicture}[
    block/.style={
        draw=#1!70!black, fill=#1!12, rounded corners=4pt,
        minimum width=2.9cm, minimum height=1.15cm, align=center, line width=0.6pt
    },
    arr/.style={-{Stealth[length=5pt, width=4pt]}, line width=0.6pt, color=gray!70!black},
    label/.style={font=\scriptsize\itshape, text=gray!60!black},
    scope/.style={font=\scriptsize\bfseries, text=black!70}
]
\node[block=blue]   (raw)  at (0,0)    {Raw NIfTI\\$\mathbf{x}_{t}$};
\node[block=orange] (n4)   at (3.4,0)  {Bias-field\\Correction};
\node[block=teal]   (tmpl) at (6.8,0)  {Within-subject\\Template};
\node[block=purple] (mask) at (10.2,0) {Brain\\Extraction};
\node[block=teal]   (mni)  at (13.6,0) {Affine to\\Standard Space};
\node[block=red]    (res)  at (17.0,0) {Single\\Resampling};
\node[block=green]  (int)  at (20.4,0) {Intensity\\Normalisation};
\draw[arr] (raw) -- (n4);
\draw[arr] (n4) -- (tmpl);
\draw[arr] (tmpl) -- (mask);
\draw[arr] (mask) -- (mni);
\draw[arr] (mni) -- (res);
\draw[arr] (res) -- (int);
\node[label] at (3.4,-0.95)  {N4};
\node[label] at (6.8,-0.95)  {rigid, iterative};
\node[label] at (10.2,-0.95) {ANTsPyNet};
\node[label] at (13.6,-0.95) {MNI152NLin2009cAsym};
\node[label] at (17.0,-0.95) {$128^3$ at 1.6\,mm};
\node[label] at (20.4,-0.95) {percentiles, $[0,1]$};
\node[scope] at (1.7,1.05)   {each scan};
\node[scope] at (10.2,1.05)  {once per participant};
\node[scope] at (18.7,1.05)  {each scan};
\draw[black!35, line width=0.5pt] (-1.45,0.85) -- (4.85,0.85);
\draw[black!35, line width=0.5pt] (5.35,0.85) -- (15.05,0.85);
\draw[black!35, line width=0.5pt] (15.55,0.85) -- (21.85,0.85);
\end{tikzpicture}%
}
\caption{Overview of the longitudinal preprocessing pipeline. The first and last stages run on every scan; the three middle stages run once per participant, on a template built from all of that participant's scans.}
\label{fig:preprocessing}
\end{figure}

\begin{algorithm}[H]
\caption{Longitudinal MRI Preprocessing Pipeline}\label{alg:preprocessing}
\begin{algorithmic}[1]
\Require Raw T1-weighted scans $\{\mathbf{x}_1, \dots, \mathbf{x}_n\}$ of one participant; template $\mathbf{T}$ with brain mask $\mathbf{M}_{\mathbf{T}}$; output grid $\Omega$ ($128^3$ voxels, 1.6\,mm)
\Ensure Preprocessed volumes $\{\mathbf{x}'_1, \dots, \mathbf{x}'_n\}$ on $\Omega$, intensities in $[0,1]$

\Statex \hspace{-\algorithmicindent}Step 1: Bias-field correction
\For{$t = 1$ to $n$}
    \State $\mathbf{b}_t \gets \textsc{N4}(\mathbf{x}_t)$ \Comment{In native space}
\EndFor

\Statex \hspace{-\algorithmicindent}Step 2: Within-subject template
\State $\mathbf{S} \gets \mathbf{b}_1$
\For{$k = 1$ to $3$}
    \State $\mathbf{R}_t \gets \textsc{RigidRegister}(\mathbf{S}, \mathbf{b}_t)$ for all $t$ \Comment{6 degrees of freedom}
    \State $\mathbf{S} \gets \frac{1}{n}\sum_{t} \mathbf{b}_t \circ \mathbf{R}_t$ \Comment{Voxel-wise mean}
\EndFor

\Statex \hspace{-\algorithmicindent}Step 3: Brain extraction, once per participant
\State $\mathbf{M} \gets \mathbbm{1}[\textsc{BrainExtract}(\mathbf{S}) > 0.5]$
\State Refine every $\mathbf{R}_t$ with the registration restricted to $\mathbf{M}$; recompute $\mathbf{S}$ and $\mathbf{M}$

\Statex \hspace{-\algorithmicindent}Step 4: Registration to standard space
\State $\mathbf{A} \gets \textsc{AffineRegister}(\mathbf{T} \odot \mathbf{M}_{\mathbf{T}},\; \mathbf{S} \odot \mathbf{M})$ \Comment{12 degrees of freedom}

\Statex \hspace{-\algorithmicindent}Step 5: Single resampling
\For{$t = 1$ to $n$}
    \State $\mathbf{y}_t \gets \mathbf{b}_t \circ (\mathbf{R}_t \circ \mathbf{A})$ evaluated on $\Omega$ \Comment{One interpolation}
\EndFor
\State $\mathbf{M}_\Omega \gets \mathbf{M} \circ \mathbf{A}$ evaluated on $\Omega$ \Comment{One mask for all timepoints}

\Statex \hspace{-\algorithmicindent}Step 6: Intensity normalisation
\For{$t = 1$ to $n$}
    \State $p_1, p_{99} \gets$ 1st and 99th percentiles of $\mathbf{y}_t$ inside $\mathbf{M}_\Omega$
    \State $\mathbf{x}'_t \gets \mathbf{M}_\Omega \odot \textsc{Clip}\big((\mathbf{y}_t - p_1)/(p_{99} - p_1),\; 0,\; 1\big)$
\EndFor
\end{algorithmic}
\end{algorithm}

\subsubsection{Step 1: Bias-Field Correction}
\label{sec:n4}

Scanner and coil effects produce a smooth variation in intensity across the image that is unrelated to tissue. We removed it with N4 bias-field correction~\cite{tustison2010n4itk}, applied to each scan in its native space within a foreground mask obtained by intensity thresholding. Before correction, intensities above the 99.9th percentile were clipped, so that isolated bright voxels do not influence the fit. The downloaded images are the original ADNI acquisitions, so this is the only inhomogeneity correction they receive.

\subsubsection{Step 2: Within-Subject Template}
\label{sec:spatial_norm}

For each participant we built a template from all of that participant's bias-corrected scans. Let $\mathbf{b}_1, \dots, \mathbf{b}_n$ be the scans and let the first scan be the initial template $\mathbf{S}^{(0)}$. At iteration $k$, every scan is registered to the current template with a rigid transformation $\mathbf{R}_t^{(k)}$ (three rotations and three translations), and the template is replaced by the voxel-wise mean of the registered scans:
\begin{equation}
    \mathbf{S}^{(k+1)} = \frac{1}{n} \sum_{t=1}^{n} \mathbf{b}_t \circ \mathbf{R}_t^{(k)}.
    \label{eq:subject_template}
\end{equation}
We used three iterations. Because the template is an average in which every scan, including the first, has been registered and resampled, no single timepoint is favoured, which is the property that makes within-subject templates unbiased~\cite{reuter2012within}. The transformation is rigid because the aim is to remove head position only: a participant's skull does not change between visits, so a rigid alignment brings the scans into register while leaving any change in brain tissue untouched.

After the brain mask of Step~3 is available, each rigid transformation is refined once more with the registration restricted to that mask, so that the neck, jaw and other non-brain tissue cannot influence the alignment of the brain. The template and its mask are then recomputed.

\subsubsection{Step 3: Brain Extraction}
\label{sec:brain_extraction}

Non-brain tissue was removed with the deep-learning brain extraction model of ANTsPyNet~\cite{tustison2021antsx}, a pretrained 3D U-Net that returns a voxel-wise brain probability $\mathbf{p}$. The binary mask is
\begin{equation}
    \mathbf{M}(i,j,k) = \mathbbm{1}\big[\mathbf{p}(i,j,k) > 0.5\big].
    \label{eq:brain_prob}
\end{equation}
The model is applied once per participant, to the within-subject template, and not to each scan. The template has a higher signal-to-noise ratio than any single scan, and one mask shared by all timepoints removes a source of spurious change. Two masks estimated separately would differ slightly at the brain boundary, and a model comparing the timepoints could read that difference as atrophy.

\subsubsection{Step 4: Registration to Standard Space}
\label{sec:mni_registration}

So that the same anatomy sits in the same place for every participant, the brain of each within-subject template was registered to the brain of the MNI152NLin2009cAsym template at 1\,mm resolution~\cite{fonov2011unbiased}, obtained with its published brain mask from TemplateFlow~\cite{ciric2022templateflow}. The registration is a 12-parameter affine transformation $\mathbf{A}$:
\begin{equation}
    \mathbf{A} = \arg\min_{\mathbf{A}'} \; \mathcal{L}_{\text{sim}}\big(\mathbf{T} \odot \mathbf{M}_{\mathbf{T}},\; (\mathbf{S} \odot \mathbf{M}) \circ \mathbf{A}'\big),
    \label{eq:registration}
\end{equation}
where $\mathbf{T}$ and $\mathbf{M}_{\mathbf{T}}$ are the template and its mask, $\mathbf{S}$ and $\mathbf{M}$ are the participant's template and mask, and $\mathcal{L}_{\text{sim}}$ is a mutual-information similarity metric.

The choice of an affine over a nonlinear registration is deliberate. A nonlinear warp reshapes each brain locally to match the template, and in doing so removes part of the regional differences in shape, including atrophy, that the model is meant to detect. An affine transformation removes only position, orientation and overall size. Because one transformation is estimated per participant and applied to all of that participant's scans, it cannot introduce a difference between timepoints.

An affine registration started from the centres of mass can settle on a poor solution when the head is strongly pitched in the scanner. We therefore estimated the transformation twice, once from the centres of mass and once from the best of a coarse search over initial rotations. When the better of the two gave a Dice overlap below 0.92 between the participant's brain mask and the template's brain mask, we added a third, staged estimate, and in every case kept the estimate with the highest overlap.

\subsubsection{Step 5: Single Resampling onto the Output Grid}
\label{sec:resampling}

The rigid transformation of each scan and the affine transformation of its participant are composed and applied in one step, so that each scan is interpolated once, directly from its bias-corrected native image:
\begin{equation}
    \mathbf{y}_t = \mathbf{b}_t \circ \big(\mathbf{R}_t \circ \mathbf{A}\big) \quad \text{evaluated on } \Omega.
    \label{eq:resample}
\end{equation}
Interpolation is linear, and the brain mask is carried to the same grid with nearest-neighbour interpolation.

The output grid $\Omega$ has $128 \times 128 \times 128$ voxels of 1.6\,mm, a field of view of 204.8\,mm, and is centred on the bounding box of the template brain. The template brain measures $144 \times 181 \times 154$\,mm, so the grid leaves a margin of about 12\,mm or more on every side and the whole brain is retained. The voxel size is a compromise: the scans are acquired at about $1.2 \times 1.0 \times 1.0$\,mm, and a grid of $128^3$ voxels at that resolution cannot contain a brain 181\,mm long. We kept the input size of the network fixed and accepted the lower resolution in return for a complete brain.

\subsubsection{Step 6: Intensity Normalisation}
\label{sec:intensity_norm}

T1-weighted intensities have no absolute scale and differ between scanners and sessions. Each resampled scan was normalised separately, using the 1st and 99th percentiles $p_1$ and $p_{99}$ of its intensities inside the brain mask:
\begin{equation}
    \mathbf{x}'_t(i,j,k) =
    \begin{cases}
        \min\!\Big(1,\; \max\!\Big(0,\; \dfrac{\mathbf{y}_t(i,j,k) - p_1}{p_{99} - p_1}\Big)\Big) & \text{if } (i,j,k) \in \mathbf{M}_\Omega, \\[8pt]
        0 & \text{otherwise.}
    \end{cases}
    \label{eq:intensity_norm}
\end{equation}
Percentiles are used in place of the minimum and maximum because a single bright voxel would otherwise set the scale for the whole volume. No spatial smoothing is applied at any stage of preprocessing.

\begin{figure}[H]
\centering
\includegraphics[width=0.86\textwidth]{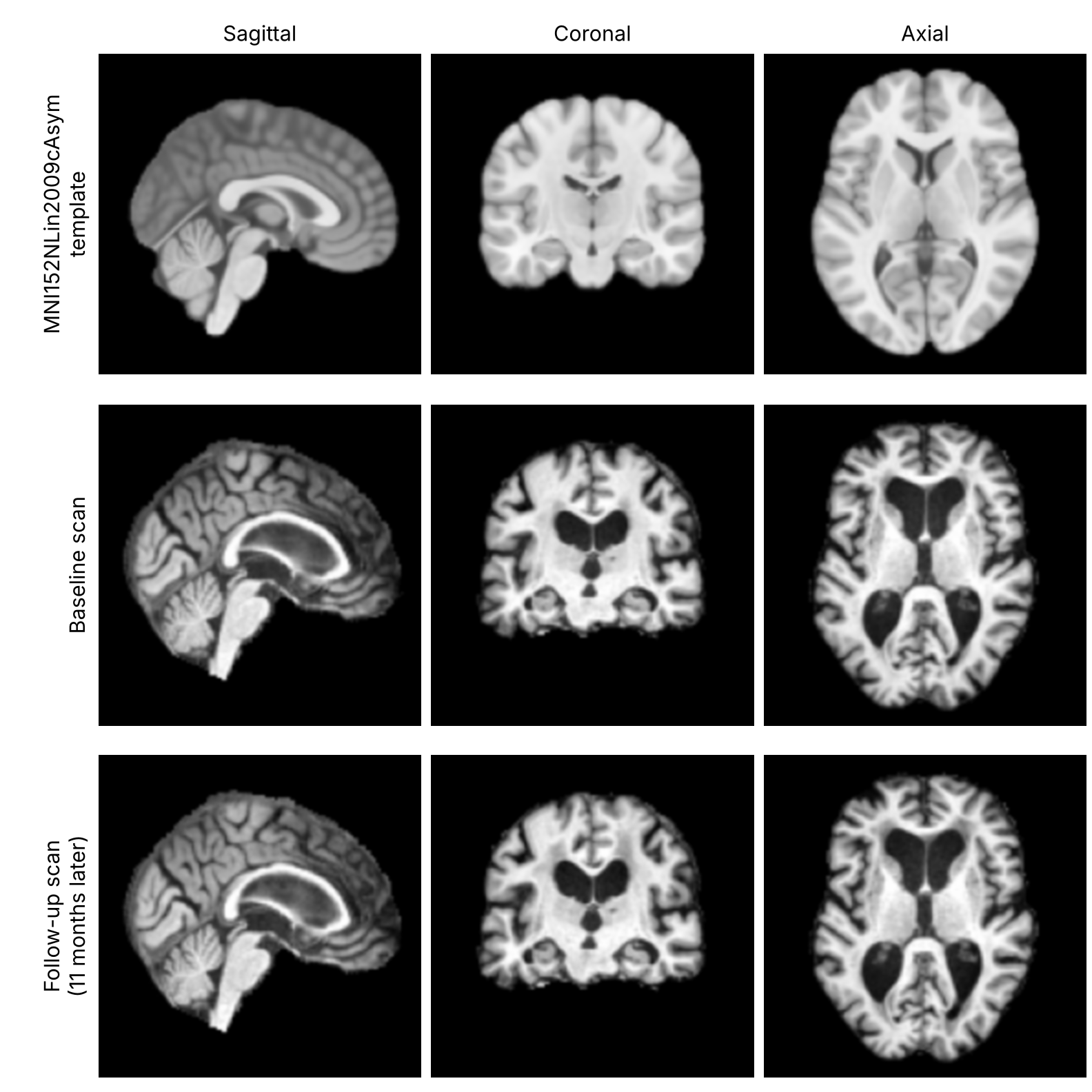}
\caption{Output of the preprocessing pipeline for one participant (ADNI 002\_S\_1268). Top row: the MNI152NLin2009cAsym template on the output grid. Middle row: baseline scan. Bottom row: follow-up scan 11 months later. Columns are the central sagittal, coronal and axial slices of the $128^3$ grid at 1.6\,mm.}
\label{fig:preprocessing_visual}
\end{figure}

Figure~\ref{fig:preprocessing_visual} shows the output for one participant of the longitudinal cohort who later converted to dementia, with the template in the top row for comparison. The baseline and follow-up scans, acquired 11 months apart, occupy the same position and orientation as the template, the skull has been removed, and the whole brain lies inside the frame, including the frontal and occipital poles, the cerebellum and the brainstem. The two scans show the same slices because both are in the participant's template space: their correlation with that template inside the brain mask is 0.99 for each, and the overlap of the participant's brain mask with the template mask is 0.96. A difference between the two rows therefore reflects a change in the participant and not a change in position.

\subsection{Data Augmentation}
\label{sec:augmentation}

The training cohort is small, so we augmented it during training to reduce overfitting. Each volume, together with its follow-up where there is one, was put through a series of geometric and intensity transformations, with new random values drawn every time a sample was loaded.

The first transformation is a left--right flip applied with probability 0.5. The brain is roughly symmetric left to right, so flipping adds variety without producing anatomy that could not exist. The second is a small random affine perturbation: a rotation of up to $5^\circ$ about each axis, a change in scale of up to 5\%, and a translation of up to two voxels. Perturbations of this size imitate the small differences in head position and size that remain between sessions and participants. The third and fourth act on intensity: Gaussian noise is added with a standard deviation drawn between 0 and 0.02, and the volume is blurred with a Gaussian kernel whose standard deviation is drawn between 0 and 0.5 voxels, which imitates differences in noise level and sharpness between scanners.

For longitudinal data, one point matters more than the rest: the relationship between the two scans must survive augmentation. We therefore applied each transformation to both volumes of a pair with the same random parameters, so that a difference between the two timepoints is not an artefact of transforming them differently. No augmentation was used during validation or testing.

\subsection{Quality Control}
\label{sec:qc}

Every preprocessed scan passed through four automated checks. The first confirmed the output dimensions of $128 \times 128 \times 128$ voxels and that all values lie in $[0,1]$. The second required that no voxel of the brain mask lies on any of the six faces of the output grid, which guarantees that no brain tissue has been cut off. The third required a correlation of at least 0.90, inside the brain mask, between the resampled scan and its participant's template, which detects a failed within-subject alignment. The fourth required a Dice overlap of at least 0.85 between the participant's brain mask and the template's brain mask on the output grid, which detects a failed registration to standard space or a failed brain extraction.

The overlap of brain masks is used for the fourth check, and not the similarity of image intensities, because an individual brain with atrophy correlates only moderately with a smooth group-average template even when the two are well aligned. A scan that failed any check was inspected visually, using a picture of the template and of all the participant's scans on the output grid of the kind shown in Figure~\ref{fig:preprocessing_visual}. Of the 2{,}893 scans, from 856 participants, 2{,}884 passed all four checks. The median Dice overlap was 0.960 (minimum 0.940) and the median correlation with the participant's template was 0.989. Nine scans from four participants were flagged and inspected on orthogonal slices. Seven of them, from two participants, were flagged by the second check. In one participant a small detached fragment of the mask, outside the brain, touched a face of the grid, and in the other eight voxels at the frontal edge did; in both, the brain lay inside the grid. The other two scans, from two participants, were flagged by the third check with correlations of 0.87 and 0.88; both were correctly aligned on inspection and differed from the participant's other scans in contrast or sharpness. No scan was excluded; two of the four participants belong to the longitudinal cohort.

\section{Proposed Method}
\label{sec:proposed_method}

This section presents the Temporal Attention Fusion Network (TAFNet), a hybrid CNN-Transformer that predicts MCI-to-dementia conversion within three years from a pair of structural MRI scans. It has three stages: a Siamese 3D CNN encoder with shared weights, which extracts spatial features; a Temporal Fusion Module with three complementary branches controlled by an adaptive gate; and a classification head. Figure~\ref{fig:tafnet} shows the whole architecture and summarises one forward pass. Given a baseline and a follow-up scan, the shared encoder extracts bottleneck features from each, the Temporal Fusion Module merges them into a single temporal representation, and the classification head turns that into a conversion probability. Sections~\ref{sec:encoder}--\ref{sec:classification} describe the three stages in turn, and Section~\ref{sec:interpretability} describes the two interpretability mechanisms.

\begin{figure}[H]
\centering
\resizebox{!}{0.82\textheight}{%
\begin{tikzpicture}[
    input/.style={
        rectangle, rounded corners=4pt, draw=blue!60!black,
        fill=blue!8, text=blue!80!black, font=\small\bfseries,
        minimum width=3.6cm, minimum height=1.0cm,
        align=center, line width=0.6pt
    },
    encoder/.style={
        rectangle, rounded corners=4pt, draw=purple!60!black,
        fill=purple!8, text=purple!80!black, font=\small\bfseries,
        minimum width=3.4cm, minimum height=2.0cm,
        align=center, line width=0.6pt
    },
    branch/.style={
        rectangle, rounded corners=4pt,
        minimum width=3.0cm, minimum height=1.2cm,
        align=center, line width=0.6pt, font=\small
    },
    branchA/.style={branch, draw=orange!70!black, fill=orange!8, text=orange!80!black},
    branchB/.style={branch, draw=teal!70!black,   fill=teal!8,   text=teal!80!black},
    branchC/.style={branch, draw=blue!60!black,   fill=blue!8,   text=blue!80!black},
    attn/.style={
        rectangle, rounded corners=4pt, draw=teal!60!black,
        fill=teal!5, text=teal!80!black, font=\small,
        minimum width=5.0cm, minimum height=0.9cm,
        align=center, line width=0.6pt
    },
    gate/.style={
        rectangle, rounded corners=4pt, draw=orange!80!black,
        fill=yellow!8, text=orange!90!black, font=\small\bfseries,
        minimum width=10.0cm, minimum height=1.0cm,
        align=center, line width=0.8pt
    },
    classifier/.style={
        rectangle, rounded corners=4pt, draw=gray!60!black,
        fill=gray!6, text=gray!80!black, font=\small,
        minimum width=8.0cm, minimum height=0.9cm,
        align=center, line width=0.6pt
    },
    output/.style={
        rectangle, rounded corners=4pt, draw=green!60!black,
        fill=green!8, text=green!80!black, font=\small\bfseries,
        minimum width=3.6cm, minimum height=0.9cm,
        align=center, line width=0.6pt
    },
    container/.style={
        rectangle, rounded corners=6pt, draw=#1,
        dashed, line width=0.7pt, inner sep=8pt
    },
    arr/.style={
        -{Stealth[length=5pt, width=4pt]},
        line width=0.6pt, color=gray!70!black
    },
    dimlabel/.style={
        font=\scriptsize\itshape, text=gray!70!black
    },
    stagelabel/.style={
        font=\small\bfseries, text=#1,
        fill=white, inner xsep=6pt, inner ysep=2pt
    },
]

\node[input] (bl)  at (0,   0) {Baseline MRI ($t_0$)};
\node[input] (m12) at (7.5, 0) {Follow-up MRI ($t_{12}$)};
\node[dimlabel, above=1pt of bl]  {$(1 \times 128^3)$};
\node[dimlabel, above=1pt of m12] {$(1 \times 128^3)$};

\node[encoder] (enc_l) at (0,   -2.8) {
    3D CNN Encoder\\[2pt]
    {\scriptsize $16{\to}32{\to}64{\to}128{\to}128$}\\[-1pt]
    {\scriptsize DCCA attention}\\[-1pt]
    {\scriptsize $4{\times}$ MaxPool $2^3$}
};
\node[encoder] (enc_r) at (7.5, -2.8) {
    3D CNN Encoder\\[2pt]
    {\scriptsize $16{\to}32{\to}64{\to}128{\to}128$}\\[-1pt]
    {\scriptsize DCCA attention}\\[-1pt]
    {\scriptsize $4{\times}$ MaxPool $2^3$}
};

\draw[dashed, purple!50!black, line width=0.5pt]
    (enc_l.east) -- node[above, font=\scriptsize\bfseries, text=purple!70!black]
    {$\theta$ shared} (enc_r.west);

\begin{scope}[on background layer]
\node[container=purple!50!black, fit=(enc_l)(enc_r),
      inner sep=12pt] (enc_container) {};
\end{scope}

\node[dimlabel] (fbl)  at (0,   -4.4) {$\mathbf{f}_{\text{BL}}  \in \mathbb{R}^{128 \times 8^3}$};
\node[dimlabel] (fm12) at (7.5, -4.4) {$\mathbf{f}_{\text{M12}} \in \mathbb{R}^{128 \times 8^3}$};

\node[branchA] (diff)      at (0,    -6.4) {
    Branch A\\[1pt]
    {\scriptsize $\Delta\mathbf{f} = \mathbf{f}_{\text{M12}} - \mathbf{f}_{\text{BL}}$}
};
\node[branchB] (crossattn) at (3.75, -6.4) {
    Branch B\\[1pt]
    {\scriptsize $\mathbf{Q}{=}\mathbf{f}_{\text{BL}}\mathbf{W}_Q$}\\[-1pt]
    {\scriptsize $\mathbf{K}{=}\mathbf{f}_{\text{M12}}\mathbf{W}_K,\; \mathbf{V}{=}\mathbf{f}_{\text{M12}}\mathbf{W}_V$}
};
\node[branchC] (concat)    at (7.5,  -6.4) {
    Branch C\\[1pt]
    {\scriptsize $[\mathbf{f}_{\text{BL}} \,;\, \mathbf{f}_{\text{M12}}] \to \text{Conv}_{1^3}$}
};

\node[attn] (mha) at (3.75, -8.2) {
    Multi-Head Attention ($H{=}4$, $d_k{=}32$)\\[-1pt]
    {\scriptsize $\text{Att}(\mathbf{Q},\mathbf{K},\mathbf{V}) = \text{softmax}\!\left(\mathbf{Q}\mathbf{K}^\top / \sqrt{d_k}\right)\mathbf{V}$}
};

\node[gate] (atg) at (3.75, -10.0) {
    Adaptive Temporal Gate (ATG)\\[-1pt]
    {\scriptsize $[\alpha, \beta, \gamma] = \text{softmax}\!\big(\mathbf{W}_2 \cdot \text{ReLU}(\mathbf{W}_1 \cdot [\text{GAP}(\mathbf{f}_{\text{BL}}) \,;\, \text{GAP}(\mathbf{f}_{\text{M12}})] + \mathbf{b}_1) + \mathbf{b}_2\big)$}
};

\node[font=\small, text=gray!80!black] (fuseq) at (3.75, -11.4) {
    $\mathbf{f}_{\text{out}} = \alpha \cdot \Delta\mathbf{f} + \beta \cdot \text{Att} + \gamma \cdot \mathbf{f}_{\text{cat}} + \mathbf{f}_{\text{BL}}$\; {\scriptsize (residual)}
};

\node[dimlabel] (fout) at (3.75, -12.3) {$\mathbf{f}_{\text{out}} \in \mathbb{R}^{128 \times 8^3}$};

\begin{scope}[on background layer]
\node[container=orange!60!black,
      fit=(diff)(concat)(mha)(atg)(fuseq)(fout),
      inner sep=10pt] (tfm_container) {};
\end{scope}

\node[classifier] (cls) at (3.75, -13.8) {
    Stage 3:\; GAP $\to$ Dropout(0.3) $\to$ FC(128,64) $\to$ ReLU $\to$ Dropout(0.3) $\to$ FC(64,1) $\to$ $\sigma$
};
\node[output] (out) at (3.75, -15.2) {$P(\text{conversion})$};

\draw[arr] (bl.south)  -- (enc_l.north);
\draw[arr] (m12.south) -- (enc_r.north);

\draw[arr] (enc_l.south) -- (fbl.north);
\draw[arr] (enc_r.south) -- (fm12.north);

\draw[arr]                (fbl.south)  -- ++(0,-0.4) -| (diff.north);
\draw[arr, gray!50!black] (fbl.south)  -- ++(0,-0.4) -| (crossattn.north);
\draw[arr, gray!50!black] (fbl.south)  -- ++(0,-0.4) -| (concat.north);
\draw[arr]                (fm12.south) -- ++(0,-0.4) -| (concat.north);
\draw[arr, gray!50!black] (fm12.south) -- ++(0,-0.4) -| (crossattn.north);
\draw[arr, gray!50!black] (fm12.south) -- ++(0,-0.4) -| (diff.north);

\draw[arr] (crossattn.south) -- (mha.north);

\draw[arr] (diff.south)   -- (diff.south   |- atg.north);
\draw[arr] (mha.south)    -- (atg.north);
\draw[arr] (concat.south) -- (concat.south |- atg.north);

\draw[arr] (atg.south) -- (fuseq.north);

\draw[arr] (fout.south) -- (cls.north);
\draw[arr] (cls.south)  -- (out.north);

\node[stagelabel=purple!70!black, anchor=south]
    at (enc_container.north) {Stage 1: Siamese 3D CNN Encoder};
\node[stagelabel=orange!70!black, anchor=south]
    at (tfm_container.north) {Stage 2: Temporal Fusion Module};

\end{tikzpicture}%
}%
\caption{Architecture of TAFNet: a shared-weight Siamese encoder, the three-branch Temporal Fusion Module governed by the Adaptive Temporal Gate, and the classification head.}
\label{fig:tafnet}
\end{figure}

\subsection{Stage 1: Siamese Encoder}
\label{sec:encoder}

Let $\mathbf{x}_{\text{BL}} \in \mathbb{R}^{1 \times 128 \times 128 \times 128}$ and $\mathbf{x}_{\text{M12}} \in \mathbb{R}^{1 \times 128 \times 128 \times 128}$ be the preprocessed baseline and follow-up volumes; the subscript M12 refers to the typical interval of about twelve months. One encoder $E$ with parameters $\theta$ processes each timepoint separately:
\begin{equation}
    \mathbf{f}_{\text{BL}} = E(\mathbf{x}_{\text{BL}};\, \theta), \quad
    \mathbf{f}_{\text{M12}} = E(\mathbf{x}_{\text{M12}};\, \theta), \quad
    \mathbf{f} \in \mathbb{R}^{128 \times 8 \times 8 \times 8}.
    \label{eq:encoder}
\end{equation}

The encoder has five 3D convolutional blocks with channels $16 \to 32 \to 64 \to 128 \to 128$; Algorithm~\ref{alg:encoder} sets out what each block computes. A block applies two $3^3$ convolutions in sequence, each with batch normalisation and a LeakyReLU activation (Algorithm~\ref{alg:encoder}, lines~4--5). Each of the first four blocks is followed by a Dynamic Contextual Channel Attention (DCCA) module, which reweights the channels, and by $2^3$ max-pooling (line~9), which halves the spatial dimensions and takes the volume from $128^3$ down to $8^3$; the fifth block has neither. The DCCA module passes the features through two further $3^3$ convolutions, global average pooling (GAP) and a two-layer fully connected bottleneck to produce channel weights $\mathbf{w}^c_i$ (lines~7--8), and multiplies them into the features, which suppresses uninformative channels and strengthens the ones that carry diagnostic signal.

\begin{algorithm}[H]
\caption{Siamese 3D CNN Encoder}\label{alg:encoder}
\begin{algorithmic}[1]
\Require Input volume $\mathbf{x} \in \mathbb{R}^{1 \times 128 \times 128 \times 128}$
\Ensure Bottleneck features $\mathbf{f} \in \mathbb{R}^{128 \times 8 \times 8 \times 8}$
\Statex
\State $C \gets [16,\, 32,\, 64,\, 128,\, 128]$ \Comment{Channel progression}
\State $\mathbf{h}_0 \gets \mathbf{x}$
\For{$i = 1$ to $5$}
    \Statex \hspace{\algorithmicindent}\hspace{\algorithmicindent}\textit{\% Double convolution block}
    \State $\mathbf{h}_i \gets \text{LeakyReLU}\big(\text{BN}\big(\text{Conv3D}_{3^3}(\mathbf{h}_{i-1},\; C[i])\big)\big)$
    \State $\mathbf{h}_i \gets \text{LeakyReLU}\big(\text{BN}\big(\text{Conv3D}_{3^3}(\mathbf{h}_i,\; C[i])\big)\big)$
    \If{$i \leq 4$} \Comment{Dynamic Contextual Channel Attention, then pooling}
        \State $\mathbf{w}^c_i \gets \sigma\big(\text{FC}\big(\text{GAP}\big(\text{ReLU}(\text{Conv3D}_{3^3}(\text{ReLU}(\text{Conv3D}_{3^3}(\mathbf{h}_i))))\big)\big)\big)$
        \State $\mathbf{h}_i \gets \mathbf{h}_i \odot \mathbf{w}^c_i$ \Comment{Channel-wise recalibration}
        \State $\mathbf{h}_i \gets \text{MaxPool3D}_{2^3}(\mathbf{h}_i)$ \Comment{Downsample spatial dims by 2}
    \EndIf
\EndFor
\State $\mathbf{f} \gets \mathbf{h}_5$ \Comment{Bottleneck: $128 \times 8 \times 8 \times 8$}
\State \Return $\mathbf{f}$
\end{algorithmic}
\end{algorithm}

We share the encoder weights between the two streams because the paired cohort is small. Separate weights do better on larger cohorts, since the baseline and follow-up encoders can then specialise~\cite{martnezmucia2021deep}, but doubling the encoder parameters on a few hundred pairs invites overfitting. Sharing weights puts both volumes in the same feature space, so the difference $\mathbf{f}_{\text{M12}} - \mathbf{f}_{\text{BL}}$ reflects real change in shape rather than a difference between encoders~\cite{qiu2023multimodal3dsiamesenet}, and it halves the parameter count. Recent work on stochastic Siamese pretraining for longitudinal imaging makes the same choice: it keeps one shared encoder and introduces the difference between timepoints later, through the masking scheme and a cross-attention decoder~\cite{ouyang2025stamp}. Our Temporal Fusion Module works the same way.

\subsection{Stage 2: Temporal Fusion Module}
\label{sec:tfm}

The Temporal Fusion Module (TFM) is what sets this architecture apart from existing longitudinal designs. Those designs each pick one way of combining the two timepoints---subtraction~\cite{qiu2023multimodal3dsiamesenet}, concatenation~\cite{martnezmucia2021deep} or cross-attention~\cite{chen2024longformer}---and each way has a known weakness (Section~\ref{sec:literature_review}). The TFM uses all three in parallel and combines their outputs with a learned gate. Algorithm~\ref{alg:tfm} gives the full procedure: line~1 computes the temporal difference (Branch~A), lines~2--10 compute cross-temporal attention (Branch~B), line~11 applies concatenation followed by convolution (Branch~C), and lines~12--14 merge the three outputs through the Adaptive Temporal Gate and a residual connection. The subsections below describe each part.

\begin{algorithm}[H]
\caption{Temporal Fusion Module (TFM)}\label{alg:tfm}
\begin{algorithmic}[1]
\Require Baseline features $\mathbf{f}_{\text{BL}} \in \mathbb{R}^{128 \times 8^3}$, follow-up features $\mathbf{f}_{\text{M12}} \in \mathbb{R}^{128 \times 8^3}$
\Ensure Fused temporal features $\mathbf{f}_{\text{out}} \in \mathbb{R}^{128 \times 8^3}$

\Statex \hspace{-\algorithmicindent}Branch A: Temporal Difference
\State $\Delta\mathbf{f} \gets \mathbf{f}_{\text{M12}} - \mathbf{f}_{\text{BL}}$ \Comment{Explicit structural change}

\Statex
\Statex \hspace{-\algorithmicindent}Branch B: Cross-Temporal Attention
\State $\tilde{\mathbf{f}}_{\text{BL}} \gets \text{Reshape}(\mathbf{f}_{\text{BL}})$ \Comment{$(B, 128, 8, 8, 8) \to (B, 512, 128)$}
\State $\tilde{\mathbf{f}}_{\text{M12}} \gets \text{Reshape}(\mathbf{f}_{\text{M12}})$ \Comment{$512 = 8^3$ spatial tokens}
\State $\mathbf{Q} \gets \tilde{\mathbf{f}}_{\text{BL}} \mathbf{W}_Q$ \Comment{Queries from baseline}
\State $\mathbf{K} \gets \tilde{\mathbf{f}}_{\text{M12}} \mathbf{W}_K$;\quad $\mathbf{V} \gets \tilde{\mathbf{f}}_{\text{M12}} \mathbf{W}_V$ \Comment{Keys/values from follow-up}
\For{$h = 1$ to $H$} \Comment{$H = 4$ attention heads}
    \State $\mathbf{A}_h \gets \text{softmax}\!\big(\mathbf{Q}_h \mathbf{K}_h^\top / \sqrt{d_k}\big)$ \Comment{$d_k = 32$; $\mathbf{A}_h \in \mathbb{R}^{512 \times 512}$}
    \State $\mathbf{O}_h \gets \mathbf{A}_h \mathbf{V}_h$ 
\EndFor
\State $\text{Att} \gets \text{Reshape}\big([\mathbf{O}_1;\ldots;\mathbf{O}_H] \cdot \mathbf{W}_O\big)$ \Comment{Back to $\mathbb{R}^{128 \times 8^3}$}

\Statex
\Statex \hspace{-\algorithmicindent}Branch C: Concatenation + Convolution
\State $\mathbf{f}_{\text{cat}} \gets \text{Conv3D}_{1^3}\!\big([\mathbf{f}_{\text{BL}} \;;\; \mathbf{f}_{\text{M12}}]\big)$ \Comment{$\mathbb{R}^{256 \times 8^3} \to \mathbb{R}^{128 \times 8^3}$}

\Statex
\Statex \hspace{-\algorithmicindent}Adaptive Temporal Gate
\State $\alpha, \beta, \gamma \gets \textsc{ATG}(\mathbf{f}_{\text{BL}},\, \mathbf{f}_{\text{M12}})$ \Comment{Algorithm~\ref{alg:atg}}
\State $\mathbf{f}_{\text{fused}} \gets \alpha \cdot \Delta\mathbf{f} \;+\; \beta \cdot \text{Att} \;+\; \gamma \cdot \mathbf{f}_{\text{cat}}$ \Comment{Weighted mixture}
\State $\mathbf{f}_{\text{out}} \gets \mathbf{f}_{\text{fused}} + \mathbf{f}_{\text{BL}}$ \Comment{Baseline residual connection}
\State \Return $\mathbf{f}_{\text{out}}$
\end{algorithmic}
\end{algorithm}

\subsubsection{Branch A: Temporal Difference}

The element-wise difference (Algorithm~\ref{alg:tfm}, line~1) captures how much structure has changed between the two timepoints, and in which direction:
\begin{equation}
    \Delta\mathbf{f} = \mathbf{f}_{\text{M12}} - \mathbf{f}_{\text{BL}} \;\in\; \mathbb{R}^{128 \times 8 \times 8 \times 8}.
    \label{eq:diff}
\end{equation}
Because the shared encoder puts both volumes in the same feature space, positive and negative values in $\Delta\mathbf{f}$ correspond to regional loss and expansion. This is the easiest branch to interpret, because the feature map is simply what changed, and it matches the clinical view of neurodegeneration as progressive atrophy. Its weakness is that subtraction erases the baseline. A patient with a naturally small hippocampus and mild atrophy may be at higher risk than one with a large hippocampus and the same absolute loss, yet the two produce the same difference vector~\cite{martnezmucia2021deep}.

\subsubsection{Branch B: Cross-Temporal Attention}

To capture non-linear relationships between regions across time, the bottleneck features are reshaped into sequences of spatial tokens $\mathbf{f} \in \mathbb{R}^{B \times 512 \times 128}$. The $512 = 8^3$ spatial positions act as the sequence and the $128$ channels as the token dimension (Algorithm~\ref{alg:tfm}, lines~2--3). Multi-head cross-attention then runs with the baseline features as queries and the follow-up features as keys and values (lines~4--10):
\begin{align}
    \mathbf{Q} &= \mathbf{f}_{\text{BL}} \mathbf{W}_Q, \quad
    \mathbf{K} = \mathbf{f}_{\text{M12}} \mathbf{W}_K, \quad
    \mathbf{V} = \mathbf{f}_{\text{M12}} \mathbf{W}_V, \label{eq:qkv} \\[4pt]
    \text{Att}(\mathbf{Q}, \mathbf{K}, \mathbf{V}) &= 
    \text{softmax}\!\left(\frac{\mathbf{Q}\mathbf{K}^\top}{\sqrt{d_k}}\right)\mathbf{V}.
    \label{eq:attention}
\end{align}
There are $H = 4$ heads with head dimension $d_k = 128/4 = 32$. In the implementation the queries and keys are layer-normalised before the attention, and a dropout of 0.1 is applied to the attention weights. At the $8^3$ bottleneck resolution the attention matrix $\mathbf{A} \in \mathbb{R}^{512 \times 512}$ costs only about 1\,MB per sample, so it is affordable. Querying the baseline against the follow-up lets each position in the baseline ask every position in the follow-up how it has changed, which places local atrophy in the context of change across the whole brain. This matters because progression in the brain is a connected cascade: a change in hippocampal volume means something different when the cortex is thinning at the same time, and subtraction alone cannot express that dependency~\cite{chen2024longformer}. The attended output is reshaped back to $\mathbb{R}^{128 \times 8 \times 8 \times 8}$.

\subsubsection{Branch C: Concatenation}

The third branch keeps both timepoints in full by concatenating them along the channel axis and then learning a reduction back to the original width (Algorithm~\ref{alg:tfm}, line~11):
\begin{equation}
    \mathbf{f}_{\text{cat}} = \text{Conv3D}_{1^3}\!\big([\mathbf{f}_{\text{BL}} \;;\; \mathbf{f}_{\text{M12}}]\big) \;:\; \mathbb{R}^{256 \times 8^3} \to \mathbb{R}^{128 \times 8^3},
    \label{eq:concat}
\end{equation}
where $[\,\cdot\;;\;\cdot\,]$ denotes concatenation along the channel axis. Ocasio and Duong~\cite{martnezmucia2021deep} flattened the features before concatenating them, which destroys the spatial layout; the $1{\times}1{\times}1$ convolution used here keeps the 3D structure intact. Both timepoints are thus preserved in full, so the gate downstream can draw on baseline vulnerability as well as follow-up state when neither explicit change nor attention is enough on its own.

\subsubsection{Adaptive Temporal Gate}

The three branches give three views of the same temporal relationship: explicit change ($\Delta\mathbf{f}$), alignment between regions (Att), and both timepoints in full ($\mathbf{f}_{\text{cat}}$). Instead of fixing how much each contributes, an Adaptive Temporal Gate (ATG) learns mixture weights for each patient, as set out in Algorithm~\ref{alg:atg}. Global average pooling first compresses the two feature maps into compact descriptors (Algorithm~\ref{alg:atg}, lines~1--2). The two 128-dimensional descriptors are concatenated into a single vector $\mathbf{g} \in \mathbb{R}^{256}$ (line~3). A multilayer perceptron (MLP) with two layers and a hidden width of 64 maps that vector to three logits (lines~4--5), and a softmax turns them into the gating coefficients (line~6):
\begin{align}
    \mathbf{g} &= [\text{GAP}(\mathbf{f}_{\text{BL}}) \;;\; \text{GAP}(\mathbf{f}_{\text{M12}})] \;\in\; \mathbb{R}^{256}, \label{eq:gate_input} \\[3pt]
    [\alpha, \beta, \gamma] &= \text{softmax}\!\big(\mathbf{W}_2 \cdot \text{ReLU}(\mathbf{W}_1 \mathbf{g} + \mathbf{b}_1) + \mathbf{b}_2\big), \label{eq:gate_weights} \\[3pt]
    \mathbf{f}_{\text{fused}} &= \alpha \cdot \Delta\mathbf{f} \;+\; \beta \cdot \text{Att} \;+\; \gamma \cdot \mathbf{f}_{\text{cat}}.
    \label{eq:fusion}
\end{align}
The softmax forces $\alpha + \beta + \gamma = 1$, so each coefficient reads as the share of the mixture that its branch contributes for that patient. A residual connection then adds the baseline representation back, keeping the static anatomy available (Algorithm~\ref{alg:tfm}, line~14):
\begin{equation}
    \mathbf{f}_{\text{out}} = \mathbf{f}_{\text{fused}} + \mathbf{f}_{\text{BL}}.
    \label{eq:residual}
\end{equation}
The gate follows attention-guided feature selection~\cite{wang2025afsnet}, but selects between ways of fusing rather than between spatial regions. The whole ATG pathway (Algorithm~\ref{alg:atg}) adds only about 16.6K parameters (256${\to}$64${\to}$3), which is negligible next to the encoder. In return it gives adaptive fusion and, through the coefficients themselves, a channel intended to make individual predictions interpretable.

\begin{algorithm}[H]
\caption{Adaptive Temporal Gate (ATG)}\label{alg:atg}
\begin{algorithmic}[1]
\Require Baseline features $\mathbf{f}_{\text{BL}} \in \mathbb{R}^{128 \times 8^3}$, follow-up features $\mathbf{f}_{\text{M12}} \in \mathbb{R}^{128 \times 8^3}$
\Ensure Gate coefficients $\alpha, \beta, \gamma \in [0,1]$ with $\alpha + \beta + \gamma = 1$
\Statex
\Statex \hspace{-\algorithmicindent}\textit{\% Compress spatial dimensions to global descriptors}
\State $\mathbf{z}_{\text{BL}} \gets \text{GAP}(\mathbf{f}_{\text{BL}})$ \Comment{$\mathbb{R}^{128 \times 8^3} \to \mathbb{R}^{128}$}
\State $\mathbf{z}_{\text{M12}} \gets \text{GAP}(\mathbf{f}_{\text{M12}})$ \Comment{$\mathbb{R}^{128 \times 8^3} \to \mathbb{R}^{128}$}

\Statex
\Statex \hspace{-\algorithmicindent}\textit{\% Concatenate global descriptors from both timepoints}
\State $\mathbf{g} \gets [\mathbf{z}_{\text{BL}} \;;\; \mathbf{z}_{\text{M12}}]$ \Comment{$\mathbf{g} \in \mathbb{R}^{256}$}

\Statex
\Statex \hspace{-\algorithmicindent}\textit{\% Two-layer MLP: $256 \to 64 \to 3$}
\State $\mathbf{g}' \gets \text{ReLU}(\mathbf{W}_1 \cdot \mathbf{g} + \mathbf{b}_1)$ \Comment{Hidden layer, $\mathbf{g}' \in \mathbb{R}^{64}$}
\State $\mathbf{l} \gets \mathbf{W}_2 \cdot \mathbf{g}' + \mathbf{b}_2$ \Comment{Logits, $\mathbf{l} \in \mathbb{R}^{3}$}

\Statex
\Statex \hspace{-\algorithmicindent}\textit{\% Softmax ensures interpretable, normalised mixture}
\State $[\alpha, \beta, \gamma] \gets \text{softmax}(\mathbf{l})$ \Comment{$\alpha$: difference, $\beta$: attention, $\gamma$: concat}
\State \Return $\alpha, \beta, \gamma$
\end{algorithmic}
\end{algorithm}

\subsection{Stage 3: Classification Head}
\label{sec:classification}

Global average pooling reduces the fused output $\mathbf{f}_{\text{out}}$ to a 128-dimensional vector, which then passes through a two-layer fully connected network with dropout before each layer:
\begin{equation}
    P(\text{conversion}) = \sigma\!\big(\mathbf{W}_4 \cdot \text{ReLU}(\mathbf{W}_3 \cdot \text{GAP}(\mathbf{f}_{\text{out}}) + \mathbf{b}_3) + \mathbf{b}_4\big).
    \label{eq:classifier}
\end{equation}
The intermediate width is 64, the dropout rate is 0.3, and $\sigma$ is the sigmoid function. TAFNet has 3{,}068{,}980 parameters in total: 2{,}944{,}560 in the encoder, 116{,}099 in the Temporal Fusion Module (16{,}643 of them in the Adaptive Temporal Gate) and 8{,}321 in the classification head. The pretrained encoder is frozen, so only the remaining 124{,}420 parameters are trained in the longitudinal phase.

Training uses a binary cross-entropy loss in which the positive class is weighted by the ratio of negative to positive pairs in the training fold, and the AdamW optimiser with a learning rate of $5 \times 10^{-4}$ and a weight decay of $10^{-4}$. The learning rate follows cosine annealing with warm restarts, with a first cycle of 10 epochs that doubles at each restart and a minimum of $10^{-6}$. Each model is trained for up to 50 epochs with a batch size of 4, gradient clipping at a norm of 1.0 and mixed precision. Training stops early when the validation area under the receiver operating characteristic curve (AUC) has not improved for 15 epochs, and the weights from the best validation epoch are kept. The random seed is 42 throughout. The models were implemented in PyTorch 2.9.1, with TorchIO 1.2.1 for augmentation, and trained on one NVIDIA L4 GPU.

\subsection{Interpretability}
\label{sec:interpretability}

TAFNet provides two read-outs intended to support interpretation, both obtained at inference time by the procedure in Algorithm~\ref{alg:interp}.

The first uses the cross-attention matrix $\mathbf{A} \in \mathbb{R}^{512 \times 512}$ from Branch~B. Following Algorithm~\ref{alg:interp}, lines~2--7, the per-head matrices are averaged over the $H$ heads into a single $\bar{\mathbf{A}}$ (line~3). Summing over query positions then gives the total attention received at each \emph{key} position, that is, at each location in the follow-up volume (line~4). Summing the other way would tell us nothing, because each row of $\bar{\mathbf{A}}$ is a softmax over keys and therefore always sums to one. The resulting 512-dimensional vector is reshaped onto the $8^3$ bottleneck grid (line~5) and upsampled to the original MRI resolution by trilinear interpolation (line~6). Once normalised, the map can be laid over the MNI152 template to show which regions the model relies on when comparing the two timepoints.

The second uses the gate coefficients $(\alpha, \beta, \gamma)$ that the ATG produces for each patient (Algorithm~\ref{alg:interp}, lines~8--9, which calls Algorithm~\ref{alg:atg}). In principle these say which kind of evidence drives a given prediction: a high $\alpha$ would mean clear, localised atrophy that explicit difference captures well, a high $\beta$ would mean spread-out, correlated change that needs cross-attention, and a high $\gamma$ would mean the prediction needs both timepoints in full. Because the coefficients sum to one, they can be plotted as a three-part composition for each patient and compared with clinical variables such as the rate of decline in the Mini-Mental State Examination (MMSE) score or APOE genotype.

\begin{algorithm}[H]
\caption{Interpretability: Attention Map and Gate Extraction}\label{alg:interp}
\begin{algorithmic}[1]
\Require Trained TAFNet with parameters $\Theta_{\text{best}}$, test pair $(\mathbf{x}_{\text{BL}},\, \mathbf{x}_{\text{M12}})$
\Ensure Spatial attention map $\mathbf{M}_{\text{attn}} \in \mathbb{R}^{128^3}$, gate profile $(\alpha, \beta, \gamma)$

\Statex \hspace{-\algorithmicindent}\textit{\% Forward pass with intermediate storage}
\State $\mathbf{f}_{\text{BL}},\, \mathbf{f}_{\text{M12}} \gets E(\mathbf{x}_{\text{BL}};\, \theta),\; E(\mathbf{x}_{\text{M12}};\, \theta)$

\Statex
\Statex \hspace{-\algorithmicindent}(a) Cross-Attention Map Extraction
\State Compute $\mathbf{A}_h \in \mathbb{R}^{512 \times 512}$ for each head $h$ \Comment{From Algorithm~\ref{alg:tfm}, line 7}
\State $\bar{\mathbf{A}} \gets \frac{1}{H}\sum_{h=1}^{H} \mathbf{A}_h$ \Comment{Head-averaged attention}
\State $\mathbf{a} \gets \sum_{i=1}^{512} \bar{\mathbf{A}}_{i,:}$ \Comment{Attention received at each key position, $\mathbf{a} \in \mathbb{R}^{512}$}
\State $\mathbf{M}_{8^3} \gets \text{Reshape}(\mathbf{a})$ \Comment{$\mathbb{R}^{512} \to \mathbb{R}^{8 \times 8 \times 8}$}
\State $\mathbf{M}_{\text{attn}} \gets \text{Upsample}(\mathbf{M}_{8^3},\; 128^3)$ \Comment{Trilinear interpolation}
\State Normalise $\mathbf{M}_{\text{attn}}$ to $[0,1]$ for overlay on MNI152 template

\Statex
\Statex \hspace{-\algorithmicindent}(b) Gate Coefficient Extraction
\State $\alpha, \beta, \gamma \gets \textsc{ATG}(\mathbf{f}_{\text{BL}},\, \mathbf{f}_{\text{M12}})$ \Comment{Algorithm~\ref{alg:atg}}
\State Record $(\alpha, \beta, \gamma)$ alongside clinical variables for correlation analysis

\Statex
\State \Return $\mathbf{M}_{\text{attn}},\; (\alpha, \beta, \gamma)$
\end{algorithmic}
\end{algorithm}

\section{Results}
\label{sec:results}

This section reports the evaluation of TAFNet, organised by claim rather than by cohort. Section~\ref{sec:cohort_protocol} sets out the cohort, how the labels were derived, and the evaluation protocol. Section~\ref{sec:adni_results} gives the main result on ADNI: discrimination on a held-out test partition that shares no participants with any cross-validation fold, compared against two single-timepoint convolutional baselines, two longitudinal baselines and a single-scan version of TAFNet. Section~\ref{sec:longitudinal_contribution} isolates what the longitudinal information contributes by comparing the model against an otherwise identical single-timepoint version of itself. Section~\ref{sec:branch_gate_ablation} varies the design choices inside the fusion module: the gate, the baseline residual and the branches. Section~\ref{sec:calibration_results} examines calibration and derives operating points. Section~\ref{sec:interpretability_results} tests the interpretability claim, and Section~\ref{sec:convergent} reports convergent and sensitivity analyses. External validation is outside the scope of this paper and is reported separately.

\subsection{Cohort and Evaluation Protocol}
\label{sec:cohort_protocol}

Conversion labels come from the ADNI diagnostic summary table (DXSUM), not from enrolment category. A participant counts as a converter if a dementia diagnosis is recorded within 36 months of the first scan of a pair. Three consequences of this definition are worth stating, because each changes the prevalence. First, a pair whose diagnosis was recorded before the follow-up scan is dropped, since the outcome would already be visible at the moment of prediction. Second, censored negatives---participants followed for too short a time to establish that they did not convert---are excluded rather than counted as stable. Third, any recorded dementia counts as conversion, whatever its cause. That third point is why the task is called MCI-to-dementia throughout rather than MCI-to-AD: of the 133 positive pairs, 125 have an Alzheimer's cause and 8 a non-Alzheimer's dementia diagnosis.

The timing of a pair is therefore as follows. The two scans are 6 to 24 months apart (183 to 730 days, median 359 days; 508 of the 604 pairs are 13 months apart or less). Every participant has an MCI diagnosis at baseline and no recorded dementia diagnosis by the second scan. The outcome window closes 36 months after the first scan. Because the model sees both scans, it predicts conversion over the 12 to 30 months that follow the second scan, which is about 24 months for a typical pair (Figure~\ref{fig:cohort}(b)).

These criteria yield 604 scan pairs from 325 participants, of which 133 pairs (22.0\%) are positive. Counted by participant, 102 of 325 (31.4\%) convert and 223 remain stable. Fourteen participants contribute one positive and one negative pair, which follows from anchoring the 36-month window to each pair rather than to the participant. Prevalence by pair and prevalence by participant are reported separately throughout, because they are not the same quantity. In total 365 candidate pairs were excluded: 228 censored, 112 with a non-MCI diagnosis at baseline, and 25 whose diagnosis preceded the follow-up scan. Figure~\ref{fig:cohort} summarises the result.

\begin{figure}[H]
\centering
\begin{tikzpicture}[
  font=\small,
  box/.style={draw=black!35, rounded corners=2pt, align=center, inner sep=5pt, minimum width=4.6cm},
  exc/.style={draw=black!25, rounded corners=2pt, align=left, inner sep=5pt, font=\footnotesize, text=black!70},
  ar/.style={-{Latex[length=4pt]}, draw=black!45}]
\node[box] (cand) {Candidate longitudinal pairs\\(MCI baseline, 6--24 month interval)\\\textbf{969 pairs}};
\node[exc, right=1.5cm of cand, yshift=-1.15cm] (drop)
  {Excluded (365 pairs):\\[2pt]
   $\bullet$ 228 censored --- insufficient follow-up\\
   $\bullet$ 112 non-MCI baseline diagnosis\\
   $\bullet$ \phantom{0}25 diagnosis precedes follow-up scan};
\node[box, below=2.3cm of cand] (final) {Analysis cohort\\\textbf{604 pairs from 325 participants}\\133 positive pairs (22.0\%)};
\node[box, below left=1.5cm and -2.3cm of final] (cvset) {Cross-validation\\519 pairs / 276 participants\\113 positive};
\node[box, below right=1.5cm and -2.3cm of final] (test) {Held-out test\\85 pairs / 49 participants\\20 positive};
\draw[ar] (cand) -- (final);
\draw[ar] ($(cand)!0.5!(final)$) -- (drop.west);
\draw[ar] (final.south) -- ++(0,-0.35) -| (cvset.north);
\draw[ar] (final.south) -- ++(0,-0.35) -| (test.north);
\end{tikzpicture}

\vspace{1mm}{\small(a) Cohort derivation}\vspace{6mm}

\begin{tikzpicture}[font=\small, x=0.33cm, y=1cm]
\draw[-{Latex[length=4pt]}] (0,0) -- (39,0) node[right] {months};
\foreach \m in {0,6,12,18,24,30,36} \draw (\m,0.12) -- (\m,-0.12) node[below] {\m};
\fill (0,0) circle (2.5pt);
\node[anchor=south, align=center] at (0,0.2) {Scan 1\\(MCI)};
\draw[line width=3pt, opacity=0.35] (6,0.45) -- (24,0.45);
\fill (12,0.45) circle (2.5pt);
\node[anchor=south west] at (5.6,0.55) {Scan 2: 6--24 months after scan 1 (median 12)};
\draw[dashed] (36,0.12) -- (36,1.0);
\draw[decorate, decoration={brace, mirror, amplitude=5pt}] (0,-0.75) -- (36,-0.75) node[midway, below=6pt] {Outcome window: dementia diagnosed within 36 months of scan 1};
\draw[decorate, decoration={brace, mirror, amplitude=5pt}] (12,-1.75) -- (36,-1.75) node[midway, below=6pt] {Horizon after scan 2: 12--30 months (24 for a typical pair)};
\end{tikzpicture}

\vspace{1mm}{\small(b) Timing of a scan pair and its outcome window}
\caption{Derivation of the analysis cohort and the evaluation partitions (a), and the timing of a scan pair relative to the outcome window (b).}
\label{fig:cohort}
\end{figure}

A held-out test partition of 15\% is set aside first, split by participant, and is never used in training, in choosing epochs, or in developing the model. It holds 85 pairs from 49 participants, 20 of them positive. Five-fold cross-validation then runs on the remaining 519 pairs from 276 participants, again split by participant so that no one's scans fall on both sides of a fold boundary. This is not a formality: 14 participants contribute a pair of each class, and splitting by pair would leak them across the partition boundary. An assertion in the loading code checks that no participant appears in two partitions. Because the folds are balanced on participants rather than pairs, and participants contribute different numbers of pairs, the folds are not equal in size and the 519 pairs do not divide evenly by five. Training partitions hold 409--423 pairs (88--93 converters) and validation partitions 96--110 pairs (20--25 converters).

The Siamese encoder is pretrained on 1,948 cross-sectional volumes from 531 participants (315 cognitively normal, 216 dementia), with the baseline diagnosis again taken from DXSUM. None of these participants appear in the 325-participant longitudinal cohort; this is asserted when the pool is generated and checked again when the data are loaded. For pretraining, the encoder is followed by the classification head of Section~\ref{sec:classification} and trained to separate the two groups, with the participants split 80/20 into training and validation sets (1{,}559 and 389 scans) and the training procedure of that section at a learning rate of $10^{-4}$. Pretraining reached a validation AUC of 0.968. The encoder is then frozen, so the longitudinal phase trains only 124,420 parameters; the full parameter breakdown is given in Section~\ref{sec:classification}.

The primary endpoint is the AUC of the five-fold ensemble on the held-out partition. Cross-validation means are secondary, and we state the gap between the two rather than hiding it. Each fold picks its best epoch on the same data it is then scored on, so the cross-validation figure is optimistic; measuring that optimism is more useful than suppressing it. The sensitivity, specificity and F1 columns of Tables~\ref{tab:main_results} and~\ref{tab:heldout} use a fixed 0.5 decision threshold and are included only for completeness. At 22\% prevalence that threshold is not a sensible operating point, and Section~\ref{sec:calibration_results} derives thresholds from the validation folds instead. AUC does not depend on a threshold, so it needs no such treatment.

\subsection{Primary Evaluation on ADNI}
\label{sec:adni_results}

Table~\ref{tab:main_results} gives cross-validation performance for all six methods and Table~\ref{tab:heldout} the matching held-out results. Two single-timepoint convolutional baselines are included, ResNet3D-18 and DenseNet3D-121, each trained end to end on the baseline scan alone. ResNet3D-18 is a 3D residual network with eight basic blocks (33.2 million parameters). DenseNet3D-121 uses the 6-12-24-16 block layout of DenseNet-121 with the growth rate reduced to 16 to limit memory use (3.0 million parameters). They show what can be achieved with no temporal information at all, and they are the floor against which every longitudinal claim in this paper is measured. The two longitudinal baselines use the same pretrained, frozen encoder as TAFNet. Siamese-Subtract subtracts the baseline feature map from the follow-up feature map and passes the result to the same classification head, so it trains 8{,}321 parameters. CNN-LSTM pools each feature map to a 128-dimensional vector, passes the two vectors in order through a one-layer, one-directional LSTM with a hidden size of 64, and classifies its final hidden state with a two-layer head, which gives 51{,}777 trainable parameters. All six methods are trained with the procedure of Section~\ref{sec:classification} and the augmentation of Section~\ref{sec:augmentation}.

\begin{table}[H]
\centering
\setlength{\tabcolsep}{5pt}
\caption{Cross-validation performance, as mean $\pm$ standard deviation across five participant-level folds.}
\label{tab:main_results}
\small
\begin{tabular}{lccccc}
\toprule
\textbf{Method} & \textbf{AUC} & \textbf{Sensitivity} & \textbf{Specificity} & \textbf{F1-Score} & \textbf{Accuracy} \\
\midrule
\multicolumn{6}{l}{\emph{Single-timepoint}} \\
ResNet3D-18 & $0.725 \pm 0.055$ & $0.398 \pm 0.271$ & $0.832 \pm 0.140$ & $0.348 \pm 0.195$ & $0.743 \pm 0.060$ \\
DenseNet3D-121 & $0.752 \pm 0.033$ & $0.287 \pm 0.394$ & $0.795 \pm 0.332$ & $0.163 \pm 0.200$ & $0.683 \pm 0.179$ \\
TAFNet-InitialOnly & $0.797 \pm 0.045$ & $0.581 \pm 0.120$ & $0.885 \pm 0.035$ & $0.580 \pm 0.061$ & $0.819 \pm 0.031$ \\
\midrule
\multicolumn{6}{l}{\emph{Longitudinal}} \\
Siamese-Subtract & $0.724 \pm 0.076$ & $0.644 \pm 0.156$ & $0.678 \pm 0.252$ & $0.482 \pm 0.083$ & $0.675 \pm 0.171$ \\
CNN-LSTM & $0.820 \pm 0.037$ & $0.617 \pm 0.128$ & $0.873 \pm 0.036$ & $0.591 \pm 0.061$ & $0.817 \pm 0.031$ \\
\textbf{TAFNet (Proposed)} & $0.814 \pm 0.040$ & $0.599 \pm 0.135$ & $0.880 \pm 0.042$ & $0.586 \pm 0.063$ & $0.819 \pm 0.028$ \\
\bottomrule
\end{tabular}
\end{table}

\begin{table}[H]
\centering
\caption{Held-out test performance. Intervals are participant-clustered bootstrap percentile intervals.}
\label{tab:heldout}
\begin{tabular}{lcccc}
\toprule
\textbf{Method} & \textbf{AUC} & \textbf{95\% CI} & \textbf{Sens.} & \textbf{Spec.} \\
\midrule
ResNet3D-18 & 0.743 & [0.526, 0.918] & 0.450 & 0.954 \\
DenseNet3D-121 & 0.725 & [0.506, 0.909] & 0.000 & 1.000 \\
Siamese-Subtract & 0.769 & [0.610, 0.909] & 0.650 & 0.831 \\
TAFNet-InitialOnly & 0.827 & [0.694, 0.937] & 0.650 & 0.846 \\
CNN-LSTM & 0.859 & [0.751, 0.951] & 0.700 & 0.862 \\
\textbf{TAFNet (Proposed)} & 0.852 & [0.737, 0.949] & 0.750 & 0.862 \\
\bottomrule
\end{tabular}
\end{table}

\begin{figure}[H]
\centering
\includegraphics[width=\textwidth]{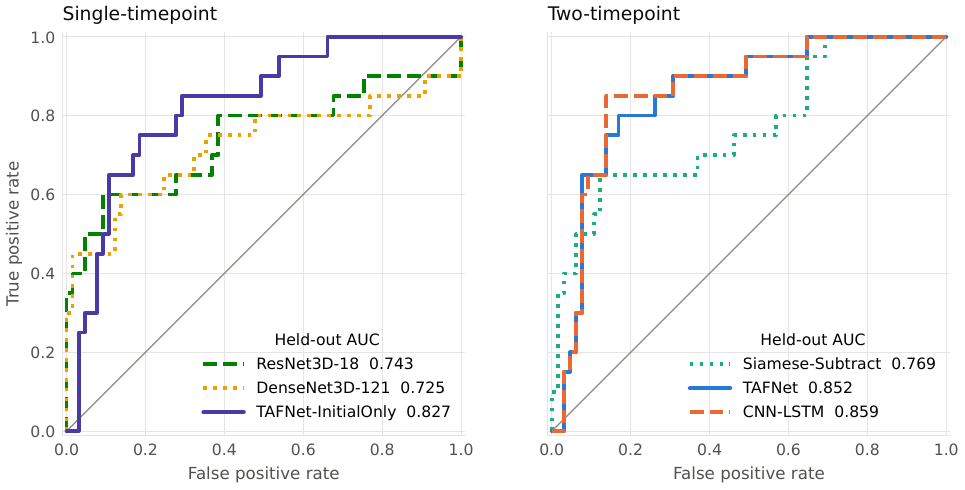}
\caption{Receiver operating characteristic curves on the held-out partition, grouped by the number of timepoints a method uses.}
\label{fig:roc}
\end{figure}

Figure~\ref{fig:roc} plots the curves, grouped by whether a method uses one timepoint or two. Three findings follow, and the evidence for them differs in strength.

The first and best supported is that the longitudinal models with learned fusion have a large and consistent advantage over the conventional single-scan baselines. The two single-timepoint baselines reach held-out AUCs of 0.743 and 0.725, while three of the four methods built on the pretrained encoder reach 0.827--0.859, and the two of them that use both timepoints reach 0.852 and 0.859; the fourth, Siamese-Subtract, is taken up below. TAFNet exceeds ResNet3D-18 by $+0.109$ AUC and DenseNet3D-121 by $+0.128$ on the held-out partition, and by $+0.089$ and $+0.062$ in the mean across cross-validation folds. The claim rests on the cross-validation partition, where 276 participants support it. On the pooled out-of-fold predictions the difference is $+0.206$ over DenseNet3D-121 (95\% CI $[+0.128, +0.286]$, $p < 0.001$) and $+0.164$ over ResNet3D-18 (95\% CI $[+0.074, +0.258]$, $p < 0.001$). The pooled differences are larger than the differences in fold means because pooling also penalises a model whose scores shift in scale from fold to fold. The two convolutional baselines do: pooled, DenseNet3D-121 reaches 0.605 and ResNet3D-18 0.648, against fold means of 0.752 and 0.725. We give both summaries for that reason and treat the fold means as the more conservative estimate of the size of the advantage. The held-out margins have the same direction, but their intervals include zero because a single-timepoint AUC estimated on 49 participants has a standard error of about 0.10. The held-out result therefore points the same way without settling the question on its own. The advantage also does not depend on which longitudinal model is used: CNN-LSTM exceeds DenseNet3D-121 by a similar margin ($+0.188$ pooled, 95\% CI $[+0.103, +0.273]$, $p < 0.001$). The gain therefore belongs to this class of model and not to one fusion design. It should not be read as the value of the second scan alone, because the longitudinal models also differ from the convolutional baselines in using the pretrained frozen encoder. Section~\ref{sec:longitudinal_contribution} separates the two.

The second finding is that learned fusion does better than plain subtraction. Siamese-Subtract uses both scans but combines them by element-wise subtraction alone. It reaches a held-out AUC of 0.769 and a cross-validation mean of 0.724, below every method with a learned fusion stage. In cross-validation TAFNet exceeds it by $+0.109$ on pooled predictions (95\% CI $[+0.038, +0.181]$, $p = 0.002$). On the held-out partition the difference is $+0.083$ and its interval includes zero (95\% CI $[-0.077, +0.250]$, $p = 0.33$), so this finding, like the first, rests on cross-validation. Simply having a follow-up scan is therefore not what produces the gain; how the two scans are combined matters. Subtraction is also the most variable method across folds (standard deviation of AUC 0.076). All timepoints of a participant are aligned to a within-subject template before they are compared (Section~\ref{sec:preprocessing}), so misregistration between the two scans is an unlikely explanation for the weakness of subtraction.

The third finding concerns the comparison with CNN-LSTM, a recurrent aggregator built on the same frozen encoder. The two models perform comparably and neither is consistently ahead. On the held-out partition TAFNet reaches 0.852 against CNN-LSTM's 0.859, a difference of $-0.007$ (95\% CI $[-0.027, +0.003]$, $p = 0.43$; DeLong $p = 0.33$). In cross-validation the mean across folds is 0.814 against 0.820, while on the pooled out-of-fold predictions TAFNet is ahead by $+0.018$ (95\% CI $[-0.003, +0.039]$, $p = 0.087$; DeLong $p = 0.047$). The sign of the difference therefore depends on the partition and on the summary, and no difference reaches significance under the clustered bootstrap. We make no claim that either is the better model. The two share the same frozen encoder and their predicted probabilities correlate at $r = 0.996$ on the held-out partition and $r = 0.981$ in cross-validation, so the paired comparison is precise even at this sample size.

We also tested equivalence, with the margin set at $\pm 0.03$ AUC, equivalence declared if the 90\% participant-clustered bootstrap interval of the difference lies inside the margin, and the pooled cross-validation predictions as the primary partition. The margin was fixed before any pooled statistic of these experiments had been computed, but after the fold-level AUCs of most models had been seen. On the held-out partition the 90\% interval is $[-0.022, +0.002]$, which lies inside the margin. On the pooled cross-validation predictions it is $[+0.001, +0.036]$, which does not, because its upper end exceeds 0.03 in TAFNet's favour. Equivalence is therefore established on the held-out partition but not on the primary one, and we describe the two models as comparable and not as equivalent. In trainable parameters the two fusion stages differ: CNN-LSTM trains 51{,}777 parameters on top of the frozen encoder and TAFNet 124{,}420.

\subsubsection*{Statistical Comparison}
\label{sec:stats}

\begin{figure}[H]
\centering
\includegraphics[width=\textwidth]{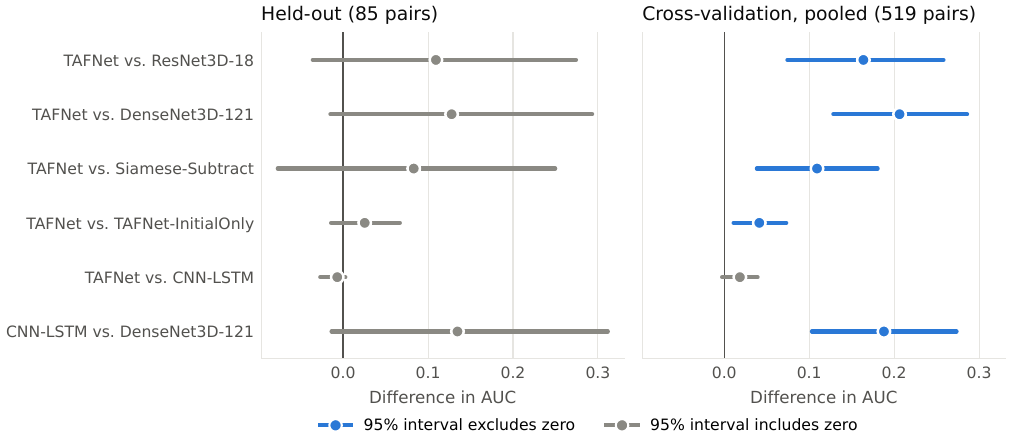}
\caption{Pairwise differences in AUC on both partitions, with participant-clustered bootstrap 95\% intervals. Intervals excluding zero are shown in colour; both panels share one horizontal scale.}
\label{fig:pairwise}
\end{figure}

The 85 held-out pairs come from only 49 participants, and two pairs from the same participant are not independent. Every interval and test in this section accounts for that. Intervals are percentile intervals from 10{,}000 bootstrap resamples in which \emph{participants}, not pairs, are drawn with replacement, and every pair belonging to a drawn participant is kept. Two-sided $p$-values come from the matching paired distribution of the difference in AUC. We also report DeLong's test for reference, but it assumes independent observations and so gives intervals that are too narrow here; wherever the two disagree, the clustered bootstrap is the one we follow. The same procedure is applied to the pooled out-of-fold cross-validation predictions (519 pairs from 276 participants). Figure~\ref{fig:pairwise} shows the six pairwise comparisons this paper relies on, on both partitions.

Two features of these comparisons shape what the rest of this paper claims. The first is that the comparisons separating the classes of method---longitudinal against single-timepoint, and learned fusion against plain subtraction---are significant on the cross-validation partition with its 276 participants, and point the same way on the 49-participant held-out partition without reaching significance there. The second is a matter of power. Methods that share the frozen encoder produce strongly correlated predictions, so comparisons among them are far better powered than the separate intervals in Table~\ref{tab:heldout} suggest. Comparisons against the independently trained single-timepoint baselines are the opposite case ($r \approx 0.71$--$0.73$ on held-out data and lower in cross-validation), and at this sample size only a large difference would reach significance. Claims of that second kind are therefore made from cross-validation throughout, with the held-out partition reported as confirmation.

For TAFNet the cross-validation mean (0.814) is $0.038$ below the held-out ensemble estimate (0.852), so the held-out figure is the higher of the two. The per-fold held-out AUCs sit close together (standard deviation 0.003), which means the ensemble adds little and the individual fold models generalise about equally well. The same direction holds for every method except DenseNet3D-121, whose cross-validation mean (0.752) is above its held-out estimate (0.725). A cross-validation mean above the held-out estimate is the expected direction for a model trained from scratch on roughly 410 pairs per fold and allowed to pick a favourable epoch on the same data it is scored on.

Two things about the single-timepoint baselines are worth recording, because they affect how Table~\ref{tab:heldout} should be read. At the 0.5 threshold the ensemble of DenseNet3D-121 calls every held-out case non-converting (specificity 1.000, sensitivity 0.000), and ResNet3D-18 reaches a specificity of 0.954 at a sensitivity of 0.450; yet both still rank cases at an AUC above 0.72. Their training curves were unstable to match, with validation AUC swinging from epoch to epoch and sensitivity falling to zero in several folds. This is how large networks behave when trained from scratch on a few hundred examples, and it is the main reason the encoder in TAFNet is pretrained and frozen.

\subsection{Longitudinal Contribution}
\label{sec:longitudinal_contribution}

Comparing TAFNet with TAFNet-InitialOnly isolates the Temporal Fusion Module, because the encoder, the classification head, the folds, the optimiser and the training schedule are all held fixed. TAFNet-InitialOnly sees only the baseline scan and nothing else differs; Table~\ref{tab:temporal} reports the comparison.

\begin{table}[H]
\centering
\caption{TAFNet with three-branch temporal fusion against the otherwise identical single-timepoint configuration. CV, cross-validation (mean $\pm$ standard deviation across folds).}
\label{tab:temporal}
\begin{tabular}{lccc}
\toprule
\textbf{Configuration} & \textbf{CV AUC} & \textbf{Held-out AUC} & \textbf{CV Sens.} \\
\midrule
TAFNet-InitialOnly & $0.797 \pm 0.045$ & 0.827 & $0.581 \pm 0.120$ \\
TAFNet (with fusion) & $0.814 \pm 0.040$ & 0.852 & $0.599 \pm 0.135$ \\
\midrule
Absolute change & $+0.017$ & $+0.025$ & $+0.018$ \\
\bottomrule
\end{tabular}
\end{table}

Adding the follow-up scan through the three-branch fusion raises AUC by $+0.017$ in the mean across cross-validation folds and by $+0.025$ on the held-out partition, and raises held-out specificity at the 0.5 threshold from 0.846 to 0.862. The gain is significant in cross-validation. On the pooled out-of-fold predictions the paired difference is $+0.041$, with a participant-clustered 95\% confidence interval of $[+0.011, +0.073]$ ($p = 0.009$; DeLong $p = 0.003$). On the held-out partition it is $+0.025$, 95\% CI $[-0.014, +0.067]$ ($p = 0.25$), the same direction with an interval that includes zero. The two versions share the encoder and their cross-validation predictions correlate at $r = 0.955$. We therefore report the contribution of temporal fusion as established in cross-validation and consistent in direction, but not significant, on the smaller held-out partition.

Two further points bound what this ablation shows.

The first is that the gain is in discrimination. At the fixed 0.5 threshold, sensitivity and F1 in cross-validation are almost unchanged by fusion (sensitivity 0.581 without and 0.599 with; F1 0.580 and 0.586), and the standard deviation of sensitivity across folds is similar (0.120 and 0.135). We therefore claim that the Temporal Fusion Module improves the ranking of cases, and not that it makes detection at a fixed threshold better or more stable.

The second is that TAFNet-InitialOnly is itself strong. It reaches a held-out AUC of 0.827 from the baseline scan alone, well above the two conventional single-timepoint baselines (0.743 and 0.725) and within 0.025 of the full model. Since it shares the encoder and head with TAFNet, what separates it from those baselines is the frozen pretrained encoder, not temporal modelling. In practice this means that much of the achievable performance is available without a second scan, which matters wherever paired scans are expensive or simply not available.

\subsection{Gate, Residual and Branch Ablations}
\label{sec:branch_gate_ablation}

We trained eight variants of TAFNet, each changing one design choice inside the Temporal Fusion Module while the pretrained encoder, the folds, the held-out partition, the optimiser and the schedule were held fixed. All eight were specified before any of them was run, and all eight are reported in Table~\ref{tab:gate_ablation}.

The first variant changes how fine-grained the gate is. As described in Section~\ref{sec:proposed_method}, the gate produces one mixture $(\alpha, \beta, \gamma)$ per participant from globally pooled descriptors. The alternative computes a mixture at each of the 512 spatial positions separately, from the three branch outputs at that position, so the fusion strategy can vary within a single volume; this raises the number of trainable parameters from 124,420 to 132,612. The second variant removes the baseline residual of Equation~\ref{eq:residual}, which adds the unfused baseline representation back to the fused output, so that the classification head works from the fused signal alone. The remaining six variants restrict the branches that enter the mixture: three keep a single branch (difference, attention or concatenation) and three leave one branch out. A branch is disabled by excluding it from the gate's softmax, so the mixture is renormalised over the branches that remain. The disabled branch receives zero weight for every participant, and the number of trainable parameters is unchanged.

\begin{table}[H]
\centering
\caption{Ablation of the Temporal Fusion Module. Each variant is compared with the full model on the same folds and the same held-out partition. Differences are variant minus full model, with 95\% confidence intervals from 10,000 participant-clustered bootstrap resamples; $p$-values are two-sided and not corrected for the eight comparisons.}
\label{tab:gate_ablation}
\footnotesize
\setlength{\tabcolsep}{4pt}
\resizebox{\textwidth}{!}{%
\begin{tabular}{lcccccc}
\toprule
 & \multicolumn{4}{c}{\textbf{Cross-validation}} & \multicolumn{2}{c}{\textbf{Held-out}} \\
\cmidrule(lr){2-5}\cmidrule(lr){6-7}
\textbf{Configuration} & \textbf{Fold mean (SD)} & \textbf{Pooled} & \textbf{Difference [95\% CI]} & $\boldsymbol{p}$ & \textbf{AUC} & \textbf{Difference [95\% CI]} \\
\midrule
Full model & 0.814 (0.040) & 0.812 & -- & -- & 0.852 & -- \\
\midrule
Per-position gate & 0.822 (0.040) & 0.819 & $+0.008$ $[-0.007, +0.024]$ & 0.30 & 0.862 & $+0.010$ $[-0.009, +0.031]$ \\
No baseline residual & 0.821 (0.040) & 0.795 & $-0.016$ $[-0.034, +0.000]$ & 0.051 & 0.856 & $+0.004$ $[-0.007, +0.016]$ \\
\midrule
Difference only & 0.819 (0.037) & 0.811 & $\phantom{+}0.000$ $[-0.014, +0.014]$ & 0.97 & 0.854 & $+0.002$ $[-0.009, +0.011]$ \\
Attention only & 0.814 (0.041) & 0.793 & $-0.018$ $[-0.037, -0.000]$ & 0.043 & 0.852 & $-0.001$ $[-0.010, +0.006]$ \\
Concatenation only & 0.819 (0.040) & 0.821 & $+0.010$ $[-0.003, +0.023]$ & 0.13 & 0.855 & $+0.002$ $[-0.009, +0.011]$ \\
\midrule
Without difference & 0.817 (0.039) & 0.802 & $-0.010$ $[-0.026, +0.007]$ & 0.24 & 0.859 & $+0.007$ $[-0.006, +0.022]$ \\
Without attention & 0.816 (0.042) & 0.808 & $-0.003$ $[-0.015, +0.009]$ & 0.59 & 0.848 & $-0.004$ $[-0.011, +0.000]$ \\
Without concatenation & 0.821 (0.042) & 0.817 & $+0.005$ $[-0.018, +0.029]$ & 0.64 & 0.853 & $+0.001$ $[-0.010, +0.009]$ \\
\bottomrule
\end{tabular}%
}
\end{table}

No variant differs reliably from the full model. On the held-out partition every variant is within 0.010 AUC of it and every interval includes zero. In cross-validation the mean across folds lies between 0.814 and 0.822 for all nine configurations, and the pooled out-of-fold differences range from $-0.018$ to $+0.010$. Predictions from each variant correlate with those of the full model at $r = 0.96$ or higher in cross-validation and $r = 0.98$ or higher on the held-out partition.

Two pooled differences approach the conventional threshold: the attention-only variant ($-0.018$, 95\% CI $[-0.037, -0.000]$, $p = 0.043$) and the variant without the baseline residual ($-0.016$, $[-0.034, +0.000]$, $p = 0.051$). We do not read either as an effect. Neither would survive correction for eight comparisons, neither appears in the mean across folds (0.814 and 0.821, against 0.814 for the full model), and neither appears on the held-out partition ($-0.001$ and $+0.004$). As in Section~\ref{sec:adni_results}, a difference that is present in the pooled figure and absent from the fold means reflects a shift in the scale of the scores between folds, not a change in ranking within them.

These results bound what can be claimed for the architecture. Any single branch performs as well as all three, removing any one branch changes nothing, and neither the granularity of the gate nor the residual path matters. The gain from the follow-up scan reported in Section~\ref{sec:longitudinal_contribution} therefore does not depend on combining three fusion strategies or on gating between them: each branch on its own already gives the classification head access to the follow-up representation, and that access appears to be what matters. These are failures to detect a difference and not demonstrations of equivalence, although every interval in Table~\ref{tab:gate_ablation} excludes a difference larger than 0.04 AUC in either direction. Section~\ref{sec:interpretability_results} examines the gate coefficients themselves and reaches a consistent conclusion.

\subsection{Calibration and Operating Point}
\label{sec:calibration_results}

The sensitivity, specificity and F1 columns of Tables~\ref{tab:main_results} and~\ref{tab:heldout} use a fixed 0.5 decision threshold, which is not a sensible operating point when only 22\% of cases are positive. At that threshold the DenseNet3D-121 ensemble reaches a specificity of 1.000 at a sensitivity of zero, and sensitivity drops to exactly zero in individual folds of both convolutional baselines. The problem is the threshold, not the models. This section replaces it.

For each fold we choose a cut-point on that fold's \emph{validation} predictions and apply it unchanged to that fold's held-out predictions. The operating point for the ensemble is the mean of the five validation-derived thresholds, applied to the mean probability across the five folds. The held-out partition is never used to choose a threshold. Candidate cut-points are the distinct predicted probabilities in the validation fold; a pair is classed as a converter when its probability is at or above the cut-point, Youden's $J$ is sensitivity $+$ specificity $- 1$, and ties are resolved in favour of the lower cut-point. We report two criteria: Youden's $J$, which we treat as primary, and the highest sensitivity that keeps validation specificity at or above 0.80, which is the constraint a clinician would recognise at this prevalence; Table~\ref{tab:operating} gives both. Intervals are participant-clustered bootstrap percentile intervals over the 49 held-out participants, as in Section~\ref{sec:stats}, and they are wide throughout because the partition holds only 20 positive pairs.

\begin{table}[H]
\centering
\caption{Held-out performance at operating points derived on the validation folds. Intervals are participant-clustered bootstrap intervals. Thr., threshold; PPV and NPV, positive and negative predictive value; BAcc, balanced accuracy.}
\label{tab:operating}
\small
\begin{tabular}{lccccccc}
\toprule
\textbf{Method} & \textbf{Thr.} & \textbf{Sens. (95\% CI)} & \textbf{Spec. (95\% CI)} & \textbf{F1} & \textbf{PPV} & \textbf{NPV} & \textbf{BAcc} \\
\midrule
\multicolumn{8}{l}{\emph{Youden's $J$ (primary)}} \\
ResNet3D-18 & 0.210 & 0.800 [0.556, 1.000] & 0.538 [0.373, 0.708] & 0.485 & 0.348 & 0.897 & 0.669 \\
DenseNet3D-121 & 0.229 & 0.750 [0.500, 0.952] & 0.615 [0.441, 0.785] & 0.500 & 0.375 & 0.889 & 0.683 \\
Siamese-Subtract & 0.462 & 0.650 [0.429, 0.870] & 0.754 [0.613, 0.877] & 0.531 & 0.448 & 0.875 & 0.702 \\
CNN-LSTM & 0.170 & 0.900 [0.750, 1.000] & 0.692 [0.527, 0.846] & 0.621 & 0.474 & 0.957 & 0.796 \\
TAFNet-InitialOnly & 0.195 & 0.850 [0.619, 1.000] & 0.662 [0.500, 0.810] & 0.576 & 0.436 & 0.935 & 0.756 \\
\textbf{TAFNet (Proposed)} & 0.239 & 0.850 [0.667, 1.000] & 0.723 [0.562, 0.869] & 0.618 & 0.486 & 0.940 & 0.787 \\
\midrule
\multicolumn{8}{l}{\emph{Validation specificity $\geq 0.80$}} \\
ResNet3D-18 & 0.350 & 0.600 [0.333, 0.857] & 0.785 [0.636, 0.917] & 0.522 & 0.462 & 0.864 & 0.692 \\
DenseNet3D-121 & 0.290 & 0.600 [0.333, 0.857] & 0.800 [0.641, 0.938] & 0.533 & 0.480 & 0.867 & 0.700 \\
Siamese-Subtract & 0.486 & 0.650 [0.429, 0.870] & 0.831 [0.698, 0.940] & 0.591 & 0.542 & 0.885 & 0.740 \\
CNN-LSTM & 0.316 & 0.850 [0.667, 1.000] & 0.754 [0.603, 0.889] & 0.642 & 0.515 & 0.942 & 0.802 \\
TAFNet-InitialOnly & 0.287 & 0.800 [0.556, 1.000] & 0.723 [0.562, 0.869] & 0.593 & 0.471 & 0.922 & 0.762 \\
\textbf{TAFNet (Proposed)} & 0.259 & 0.800 [0.556, 1.000] & 0.738 [0.588, 0.875] & 0.604 & 0.485 & 0.923 & 0.769 \\
\bottomrule
\end{tabular}
\end{table}

Three points follow from the table. First, the apparently useless behaviour of the single-timepoint baselines was an artefact of the threshold. ResNet3D-18 moves from a sensitivity of 0.450 to 0.800 and DenseNet3D-121 from zero to 0.750, at specificities of 0.538 and 0.615. No AUC changes, because AUC does not depend on a cut-point. Second, TAFNet and CNN-LSTM have the best operating-point profiles of the six and are close to each other. Under Youden's $J$ TAFNet reaches a sensitivity of 0.850 at a specificity of 0.723 (balanced accuracy 0.787, negative predictive value 0.940) and CNN-LSTM a sensitivity of 0.900 at a specificity of 0.692 (balanced accuracy 0.796, negative predictive value 0.957). Under the specificity constraint the corresponding balanced accuracies are 0.769 and 0.802. CNN-LSTM is nominally ahead on most columns, by amounts far smaller than the intervals: these numbers rest on 20 positive cases, and the sensitivity intervals span roughly $\pm 0.13$ to $\pm 0.22$. We offer them as a description of how the models behave at a sensible cut-point, not as a comparative claim. Third, TAFNet's two criteria pick similar thresholds (0.239 and 0.259), while the per-fold cut-points vary widely (standard deviations of 0.06 to 0.37 depending on the method and criterion). Variation of that size is expected at this sample size, and it is a reason to state the operating point explicitly rather than fall back on a convention.

Ranking cases correctly and assigning realistic probabilities are two different things: a model can do the first while doing the second badly. Table~\ref{tab:calibration} gives the Brier score on both partitions and the expected calibration error (ECE) on the held-out partition, along with the effect of recalibration. Recalibration maps are fitted for each fold on that fold's validation predictions, applied to that fold's held-out predictions, and then averaged. As with thresholds, no map is ever fitted on held-out data.

\begin{table}[H]
\centering
\caption{Calibration on both partitions. Lower is better throughout. ECE, expected calibration error; CV, cross-validation (pooled out-of-fold predictions). The last two columns give the held-out Brier score after recalibration.}
\label{tab:calibration}
\small
\begin{tabular}{lccccc}
\toprule
& \multicolumn{2}{c}{\textbf{Held-out}} & \textbf{CV} & \multicolumn{2}{c}{\textbf{Held-out, recalibrated}} \\
\cmidrule(lr){2-3} \cmidrule(lr){4-4} \cmidrule(lr){5-6}
\textbf{Method} & \textbf{Brier (95\% CI)} & \textbf{ECE} & \textbf{Brier} & \textbf{Platt} & \textbf{Isotonic} \\
\midrule
ResNet3D-18 & 0.150 [0.095, 0.216] & 0.103 & 0.188 & 0.153 & 0.156 \\
DenseNet3D-121 & 0.159 [0.105, 0.224] & 0.083 & 0.240 & 0.145 & 0.149 \\
Siamese-Subtract & 0.199 [0.177, 0.223] & 0.229 & 0.213 & 0.146 & 0.148 \\
CNN-LSTM & 0.129 [0.070, 0.194] & 0.081 & 0.145 & 0.122 & 0.133 \\
TAFNet-InitialOnly & 0.143 [0.083, 0.211] & 0.089 & 0.146 & 0.135 & 0.140 \\
\textbf{TAFNet (Proposed)} & 0.136 [0.070, 0.212] & 0.079 & 0.147 & 0.127 & 0.131 \\
\bottomrule
\end{tabular}
\end{table}

\begin{figure}[H]
\centering
\includegraphics[width=0.62\textwidth]{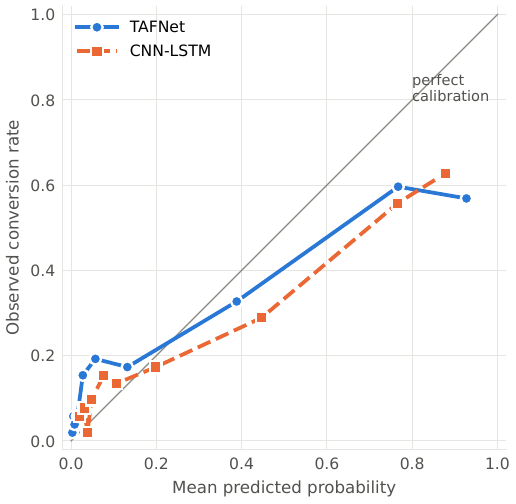}
\caption{Reliability of predicted probabilities, pooled out-of-fold, in ten equal-count bins. The diagonal marks perfect calibration.}
\label{fig:calibration}
\end{figure}

Figure~\ref{fig:calibration} shows the reliability of the pooled out-of-fold probabilities for TAFNet and CNN-LSTM. Both follow the diagonal through the middle of the range and both overestimate risk in their highest bins. In its top two deciles TAFNet predicts mean probabilities of 0.77 and 0.93 where the observed conversion rates are 0.60 and 0.57; CNN-LSTM predicts 0.77 and 0.88 where the observed rates are 0.56 and 0.63. At the low end both underestimate slightly, TAFNet more so because its lowest scores sit closer to zero.

The figures in Table~\ref{tab:calibration} should be read against the observed prevalence, which is 0.235 on the held-out partition and 0.218 in cross-validation. Expected calibration error is computed over five quantile bins, and the Platt and isotonic maps are fitted on the validation folds only. TAFNet, CNN-LSTM and TAFNet-InitialOnly have the lowest Brier scores, and calibration does not separate TAFNet from CNN-LSTM. CNN-LSTM has the lower Brier score on both partitions and TAFNet the lower expected calibration error on the held-out partition, and the paired difference in Brier score between them is small: $+0.007$ on the held-out partition (participant-clustered 95\% CI $[-0.003, +0.020]$, $p = 0.20$) and $+0.002$ in cross-validation (95\% CI $[-0.004, +0.008]$, $p = 0.55$). TAFNet's mean predicted probability on the held-out partition is 0.292 against an observed prevalence of 0.235, and its calibration slope is below one (0.57 held out, 0.44 in cross-validation), so its probabilities are more extreme than the outcomes justify. Platt scaling moves its Brier score from 0.136 to 0.127.

Siamese-Subtract is the method for which recalibration matters most. Its mean predicted probability is 0.464 against the same observed 0.235, and its expected calibration error of 0.229 is the largest of the six; isotonic regression recovers much of this, moving the Brier score from 0.199 to 0.148. TAFNet is better calibrated than Siamese-Subtract in cross-validation ($\Delta$Brier $= -0.066$, 95\% CI $[-0.096, -0.034]$, $p < 0.001$); on the held-out partition the difference is the same size but its interval includes zero ($-0.063$, 95\% CI $[-0.120, +0.005]$, $p = 0.067$). Plain subtraction is therefore both worse at ranking and worse calibrated than learned fusion in cross-validation, which supports the argument of Section~\ref{sec:adni_results}.

\subsection{Model Interpretability}
\label{sec:interpretability_results}

Section~\ref{sec:interpretability} introduced the Adaptive Temporal Gate as a way of explaining individual predictions: because $\alpha$, $\beta$ and $\gamma$ sum to one, each participant's prediction can be described as a mixture of explicit difference, cross-temporal attention and concatenation. This subsection asks whether that description tells us anything about the participant. It does not. We report the analysis in full, because a negative result of this kind limits what adaptive gating can be claimed to deliver.

Gate coefficients were extracted for every pair in both partitions using the procedure in Algorithm~\ref{alg:interp}, with the branch order \emph{(difference, attention, concatenation)} read from the implementation. Each cross-validation pair is scored by the one fold model that did not train on it; each held-out pair is scored by all five, and its coefficients are averaged across them. All intervals are percentile intervals from 2{,}000 bootstrap resamples of \emph{participants}, as in Section~\ref{sec:stats}.

\begin{figure}[H]
\centering
\includegraphics[width=\textwidth]{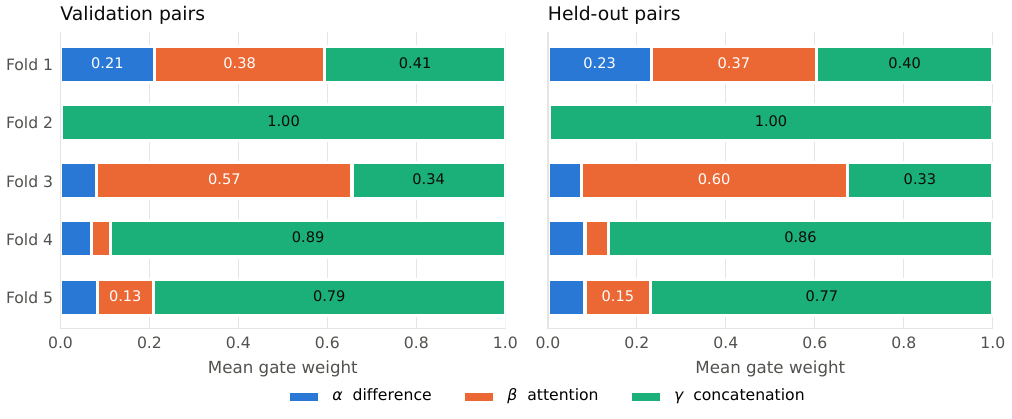}
\caption{Mean adaptive gate coefficients by fold model, on each fold's validation pairs and on the held-out partition.}
\label{fig:gate_folds}
\end{figure}

The mixture the gate learns is not stable across folds. Figure~\ref{fig:gate_folds} gives the mean coefficient each fold model assigns. On the held-out partition, the fold~2 model puts 0.997 of the weight on concatenation and effectively switches the other two branches off; the fold~4 and fold~5 models also favour concatenation (0.864 and 0.768); the fold~3 model puts most weight on cross-temporal attention (0.598); and the fold~1 model spreads the weight across all three (0.232, 0.372 and 0.396). The difference branch never dominates, and its mean never rises above 0.24. Five models trained on overlapping data under the same protocol therefore settle on qualitatively different fusion strategies.

The coefficients vary far more between folds than between participants. On the held-out partition every pair is scored by all five fold models, which lets the two sources of variation be separated. The standard deviation across folds for one pair averages 0.078 for $\alpha$, 0.227 for $\beta$ and 0.269 for $\gamma$, against 0.033, 0.025 and 0.050 across participants for the fold-averaged coefficients. The coefficients therefore describe the trained model much more than they describe the participant.

The gate largely restates the model's own output, and does so inconsistently. On the held-out partition the fold-averaged coefficients correlate with the predicted conversion probability at $\rho = +0.726$ ($[+0.455, +0.900]$) for $\alpha$, $+0.763$ ($[+0.566, +0.850]$) for $\beta$ and $-0.860$ ($[-0.943, -0.671]$) for $\gamma$. Within individual fold models, each scored on its own validation pairs, the relation ranges from near zero to near one in absolute value, and for $\beta$ it changes sign: $\rho = -0.969$ in fold~1 and $+0.907$ and $+0.926$ in folds~4 and~5. A higher attention weight thus signals lower predicted risk in one model and higher predicted risk in another.

No clinical association survives adjustment for the prediction. In cross-validation the coefficients are unrelated to the rate of MMSE change ($|\rho| \le 0.052$, $p \ge 0.38$) and to time to conversion among converters ($|\rho| \le 0.109$, $p \ge 0.28$), and their means do not differ between converters and non-converters ($p \ge 0.16$). On the held-out partition, taken on their own, all three are associated with the rate of MMSE change: $\alpha$ $-0.377$ ($[-0.570, -0.155]$), $\beta$ $-0.514$ ($[-0.673, -0.251]$) and $\gamma$ $+0.517$ ($[+0.285, +0.675]$), each with $p = 0.001$. The predicted probability itself correlates with MMSE change at $\rho = -0.608$ on this partition (Section~\ref{sec:convergent}), and adjusting for it removes all three associations: the partial rank correlations are $+0.083$ ($[-0.196, +0.274]$), $-0.091$ ($[-0.350, +0.161]$) and $+0.014$ ($[-0.204, +0.302]$). The fold-averaged coefficients also differ between converters and non-converters on this partition, unlike in cross-validation: the mean difference is $+0.027$ for $\alpha$ ($[+0.012, +0.043]$, $p = 0.002$), $+0.022$ for $\beta$ ($[+0.005, +0.036]$, $p = 0.017$) and $-0.049$ for $\gamma$ ($[-0.070, -0.025]$, $p < 0.001$). The signs are those expected from the correlation of each coefficient with the predicted probability reported above; we did not adjust this comparison separately. What looks like clinical signal in the held-out gate is the predicted risk seen a second time, made visible there because averaging over five models smooths out the fold-specific behaviour.

Together these results do not support the interpretability claim the gate was designed to make. The mechanism is part of a model that achieves the discrimination reported in Section~\ref{sec:adni_results}, but its coefficients are not a stable or independently meaningful description of a participant. We therefore make no interpretability claim for TAFNet. We do not report the spatial attention maps of Algorithm~\ref{alg:interp}(a). Upsampled from the $8^3$ bottleneck, such a map has an effective resolution of about 26\,mm, which is coarser than the structures it would be used to point at, and we could not find a way to validate these maps that was not circular.

\subsection{Convergent and Sensitivity Analyses}
\label{sec:convergent}

Three questions bear on whether the discrimination reported above reflects anything clinically real, and all three can be answered from the existing predictions without training anything new. Every correlation below is a Spearman coefficient with participant-clustered bootstrap intervals.

The first question is whether the predicted risk tracks a clinical measure the model never saw, and it does. In the cross-validation pool the predicted probability correlates with the participant's later rate of cognitive decline at $\rho = -0.324$ (95\% CI $[-0.427, -0.218]$, $n = 423$), and on the held-out partition at $\rho = -0.608$ ($[-0.747, -0.373]$, $n = 76$). Participants the model scores as high risk lose MMSE points faster. Decline rate is continuous and was never used in training, so this is stronger evidence of validity than the binary outcome alone.

\begin{figure}[H]
\centering
\includegraphics[width=\textwidth]{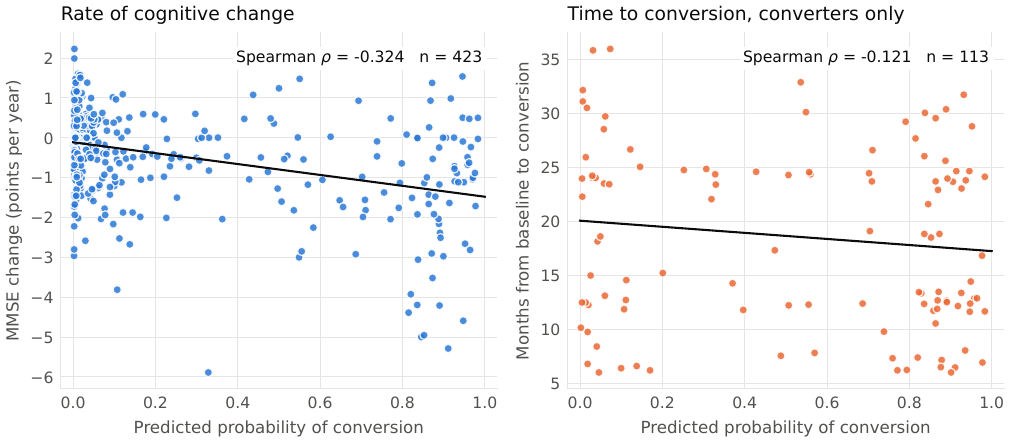}
\caption{Predicted risk against two clinical quantities the model never observed: the subsequent rate of MMSE change (left) and time to conversion among converters (right). Predictions are pooled out-of-fold.}
\label{fig:validity}
\end{figure}

The second question is whether the score also says \emph{when} conversion happens, and it does not. Among converters, the predicted probability is uncorrelated with time to diagnosis ($\rho = -0.121$, $[-0.311, +0.073]$, $n = 113$ in cross-validation; $\rho = +0.115$, $[-0.417, +0.578]$, $n = 20$ held out). The model separates converters from non-converters and orders them by severity, but says nothing usable about timing. Figure~\ref{fig:validity} shows both relationships side by side. This is a real limitation for the clinical use most often proposed for such tools, and it is the strongest argument for treating the task as time-to-event prediction rather than classification within a fixed window.

The third question is how much of this performance a single bedside test would recover. Baseline MMSE used on its own, by rank, reaches an AUC of 0.787 ($[0.737, 0.832]$) in cross-validation and 0.761 ($[0.604, 0.901]$) on held-out data, against 0.812 and 0.852 for the model. Imaging therefore adds real but modest discrimination over a test that takes minutes and needs no equipment. The two are not redundant---the model's output correlates with baseline MMSE at only $\rho = -0.38$ and $-0.21$ on the two partitions---but any claim for imaging-based prognosis should be measured against this reference rather than against chance (Table~\ref{tab:sensitivity}).

\begin{table}[H]
\centering
\caption{Convergent and sensitivity analyses. Intervals are participant-clustered bootstrap percentile intervals. The cross-validation column uses pooled out-of-fold predictions.}
\label{tab:sensitivity}
\small
\begin{tabular}{lcc}
\toprule
\textbf{Analysis} & \textbf{Cross-validation} & \textbf{Held-out} \\
\midrule
Model AUC (reference) & 0.812 & 0.852 \\
Baseline MMSE alone, AUC & 0.787 [0.737, 0.832] & 0.761 [0.604, 0.901] \\
AUC, Alzheimer's-aetiology positives only & 0.817 [0.759, 0.870] & 0.849 [0.729, 0.947] \\
\midrule
$\rho$, risk vs.\ MMSE decline rate & $-0.324$ [$-0.427$, $-0.218$] & $-0.608$ [$-0.747$, $-0.373$] \\
$\rho$, risk vs.\ time to conversion & $-0.121$ [$-0.311$, $+0.073$] & $+0.115$ [$-0.417$, $+0.578$] \\
\bottomrule
\end{tabular}
\end{table}

The outcome definition counts any recorded dementia, not Alzheimer's disease specifically. Restricting the positives to cases with an Alzheimer's cause barely changes discrimination, from 0.812 to 0.817 in cross-validation and from 0.852 to 0.849 on held-out data, so the reported performance is not an artefact of the broader definition.

We also compared TAFNet and CNN-LSTM within subgroups of the cross-validation pairs defined by acquisition, an analysis specified before these results were examined. Its purpose was to see whether either model is more robust to the heterogeneity described in Section~\ref{sec:dataset}. In the 247 pairs that combine an accelerated with a non-accelerated scan (46 positive), TAFNet reaches an AUC of 0.807 and CNN-LSTM 0.800. In the 272 pairs with consistent acceleration (67 positive) the figures are 0.813 and 0.788. Neither model therefore loses discrimination on mixed-acceleration pairs, and the difference between the two models does not depend on whether acceleration is mixed (interaction $p = 0.42$). By vendor, both models discriminate less well on pairs acquired on GE systems (124 pairs, 27 positive: TAFNet 0.725, CNN-LSTM 0.655) than on Siemens systems (300 pairs, 63 positive: 0.841 and 0.837) or Philips systems (81 pairs, 19 positive: 0.818 and 0.829). On GE pairs TAFNet is ahead of CNN-LSTM by $+0.070$ (95\% CI $[+0.018, +0.143]$); this is one of several subgroup contrasts, none corrected for multiplicity, and we treat it as a hypothesis for later work and not as a finding. The lower performance of both models on GE acquisitions, most of which use the IR-(F)SPGR sequence, is taken up in Section~\ref{sec:limitations}.

One analysis we had planned is not possible with this cohort. Breaking performance down by the interval between scans tells us nothing here, because few pairs have a long interval: 432 of the 519 cross-validation pairs are 13 months apart or less, and the two longer strata (13 to 18 and 18 to 24 months) contain only three and four positive pairs. Any interval effect in these data cannot be distinguished from sampling noise, so we report no interval-stratified result rather than one computed from a handful of events.

\section{Discussion}
\label{sec:discussion}

This study proposed a temporal attention fusion network for predicting conversion from MCI to dementia from a pair of structural MRI scans, and evaluated it under a protocol strict enough that the result can be trusted. Three findings follow, which differ in how strongly they are supported and in what they allow us to claim.

\subsection{Architecture-Independent Temporal Gain}

The clearest result is that models combining both scans through learned fusion discriminate substantially better than conventional single-scan networks. Two convolutional baselines trained on the baseline scan alone reach held-out AUCs of 0.743 and 0.725, while TAFNet reaches 0.852 and CNN-LSTM 0.859. On the cross-validation partition, where 276 participants support the comparison, TAFNet exceeds DenseNet3D-121 by $+0.062$ in the mean across folds and by $+0.206$ on pooled predictions (95\% CI $[+0.128, +0.286]$, $p < 0.001$). A longitudinal model with an unrelated architecture produces a gain of the same order ($+0.188$ pooled, $p < 0.001$), so the gain does not depend on any particular fusion design. Two different quantities must be kept apart here. The margin over the convolutional baselines separates classes of model that differ in more than the number of scans, since the longitudinal models also use the pretrained encoder. The contribution of the second scan itself, measured with everything else held fixed, is $+0.041$ AUC on pooled cross-validation predictions (95\% CI $[+0.011, +0.073]$, $p = 0.009$) and $+0.025$ on held-out data, where it does not reach significance (Section~\ref{sec:longitudinal_contribution}). The direction and the cross-validation result agree with the clinical literature showing that rates of atrophy carry prognostic information beyond cross-sectional volumes~\cite{jack2005brain,risacher2010longitudinal}.

How the two scans are combined matters as well. Siamese-Subtract takes both volumes but combines them by element-wise subtraction alone. It reaches 0.769 on held-out data and is significantly below TAFNet in cross-validation ($\Delta = +0.109$, 95\% CI $[+0.038, +0.181]$, $p = 0.002$), although not on the smaller held-out partition ($\Delta = +0.083$, $p = 0.33$). Subtraction discards the baseline and keeps only the change, whereas the learned fusion keeps both. Because all timepoints of a participant are aligned to a within-subject template before any comparison, misregistration between the two scans is an unlikely explanation for this weakness. Acquisition effects remain, however: 284 of the 604 pairs combine an accelerated with a non-accelerated acquisition and 41 change scanner model, and the within-subject alignment is rigid (Section~\ref{sec:limitations}).

\subsection{A Measurable Ceiling}

Breaking the improvement into parts tells us more than a single number. In the mean across cross-validation folds the sequence runs DenseNet3D-121 0.752 $\rightarrow$ TAFNet-InitialOnly 0.797 $\rightarrow$ TAFNet 0.814, and on the held-out partition 0.725 $\rightarrow$ 0.827 $\rightarrow$ 0.852. On both partitions roughly three-quarters of the improvement over the single-timepoint convolutional baseline comes from the frozen pretrained encoder and the architecture around it, and the remainder, $+0.017$ and $+0.025$ AUC, from the second timepoint.

Section~\ref{sec:convergent} gives a second way of reading that figure. Baseline MMSE alone, a test that takes minutes and needs no equipment, reaches an AUC of 0.787 in cross-validation and 0.761 on held-out data, against the model's 0.812 and 0.852. The imaging pipeline---two registered volumes, a pretrained 3D encoder and a learned fusion stage---therefore yields a real but limited improvement over a bedside test, and the two are only partly redundant. Prognostic claims for imaging should be measured against that reference rather than against chance.

Much of the prognostic signal available from two T1-weighted volumes is therefore already in the first one, and the follow-up scan adds a gain that is established in cross-validation but modest in size. This suggests that further architectural work on fusing paired T1 scans has limited room, and that larger gains are more likely to come from extra modalities or from longer and more frequent observation.

\subsection{Comparable Performance to a Recurrent Baseline}

TAFNet and CNN-LSTM perform comparably. On the held-out partition the difference is $-0.007$ AUC (95\% CI $[-0.027, +0.003]$, $p = 0.43$), and on pooled cross-validation predictions it is $+0.018$ (95\% CI $[-0.003, +0.039]$, $p = 0.087$); the mean across folds slightly favours CNN-LSTM (0.820 against 0.814). The comparison is precise, not inconclusive: the two share a frozen encoder and their predictions correlate at $r = 0.98$ or above. The equivalence test with a margin of $\pm 0.03$ AUC is met on the held-out partition but not on the pooled cross-validation predictions, where the interval extends to $+0.036$ in TAFNet's favour (Section~\ref{sec:adni_results}). We therefore claim neither that TAFNet is better nor that the two are equivalent. Calibration and operating points give the same picture: the two models are close on every measure, with neither consistently ahead (Section~\ref{sec:calibration_results}).

The breakdown above suggests why. Suppose most of the recoverable signal is the rate of atrophy in disease-relevant regions, and suppose that quantity is fixed mainly by the encoder's representation rather than by whatever operator is applied across time. Any competent way of combining two encoder outputs should then arrive at roughly the same estimate. A three-branch gated mixture and a recurrent layer would be two routes to the same ceiling. On that reading the similarity is not a fault in either model but a property of the task when only two timepoints are available. We offer this as an interpretation consistent with our own decomposition, not as a demonstrated mechanism. Demonstrating it would mean relating model outputs to regional atrophy measured independently, which the present imaging-only design cannot do.

We find no measured advantage of TAFNet over CNN-LSTM, and we do not claim one. Discrimination, calibration and operating points do not separate the two. TAFNet is not the smaller model: its fusion stage trains 124{,}420 parameters against 51{,}777 for the recurrent layer and its classifier. The branch ablations show that the three-branch design is not what produces its accuracy (Section~\ref{sec:branch_gate_ablation}), and the gate does not explain individual predictions (Section~\ref{sec:interpretability_results}). Computational cost does not separate them either. On the same hardware (one NVIDIA L4), a training epoch takes 87\,s for both models, and a forward pass for one pair of scans takes 74.5\,ms for TAFNet and 74.0\,ms for CNN-LSTM at full precision. The shared encoder accounts for more than 98\% of that time; the stage after it takes 1.1\,ms and 0.5\,ms. The one place where TAFNet is nominally ahead, pairs acquired on GE scanners (Section~\ref{sec:convergent}), is a single uncorrected contrast among eight and is a hypothesis for future work, not a result. TAFNet is therefore best read as a comparable alternative to a recurrent baseline. What the evidence does support is narrower and does not depend on the choice between them: models that see two scans outperform models that see one, and learned fusion of the two representations outperforms subtracting them. Both differences are significant in cross-validation and have the same direction, without reaching significance, on the smaller held-out partition.

\subsection{The Limits of Adaptive Gating}

The adaptive gate was meant to explain individual predictions, and Section~\ref{sec:interpretability_results} shows it does not. The mixture differs qualitatively between fold models trained on overlapping data, and the coefficients vary several times more between folds than between participants. The direction in which a coefficient relates to predicted risk changes from one fold model to another. Where the coefficients do relate to cognitive decline, on the held-out partition, the relation disappears once the model's own predicted probability is adjusted for.

Section~\ref{sec:branch_gate_ablation} closes the argument from the other side. Making the gate finer-grained changes nothing, removing the baseline residual changes nothing, and restricting the mixture to any one branch or any two changes nothing. That last result also explains the instability: when every branch on its own supports the same discrimination, nothing in the training signal favours one mixture over another, and each fold model is free to settle on a different one. A mechanism whose coefficients do not describe the participant, whose parameterisation does not affect the result, and whose branches are interchangeable is not what carries the model's performance.

We report this because the pattern is likely to appear in other architectures too. Mixture weights that sum to one and are computed from the same features that produce the prediction will tend to repeat that prediction. A plain correlation between such weights and a clinical variable therefore tells us very little. Showing that a gating mechanism is interpretable means showing that it relates to something external \emph{after} conditioning on the model's output, and that it is stable across independently trained models. We suggest both checks should be routine wherever learned gating is offered as an explanation.

\subsection{Thresholds and Calibration}

One practical finding has immediate consequences: the usual 0.5 decision threshold makes no sense at this prevalence. At 0.5, the DenseNet3D-121 ensemble calls every held-out case non-converting, with sensitivity also falling to zero in individual folds of ResNet3D-18. The models rank cases perfectly well while appearing clinically useless. Choosing cut-points on the validation folds and applying them unchanged to held-out data raises TAFNet's sensitivity from 0.750 to 0.850 at a specificity of 0.723, with a negative predictive value of 0.940 and balanced accuracy of 0.787. Sensitivity and specificity reported at a default threshold should therefore be read as a property of that threshold, not of the model.

\begin{figure}[H]
\centering
\includegraphics[width=0.72\textwidth]{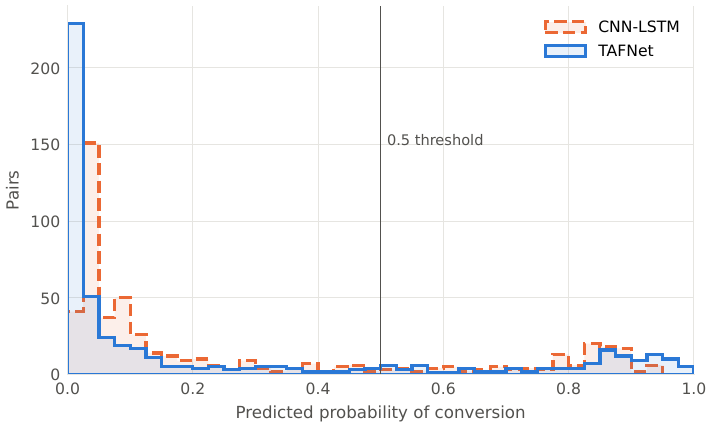}
\caption{Distribution of predicted conversion probabilities, pooled out-of-fold.}
\label{fig:scores}
\end{figure}

Calibration makes a related point. TAFNet's Brier score on the held-out partition is 0.136 and its expected calibration error 0.079, close to CNN-LSTM's 0.129 and 0.081, but its scores are pushed towards the extremes. The held-out median is 0.086 and 25.9\% of its predictions fall below 0.01, against a median of 0.149 and no predictions below 0.01 for CNN-LSTM. The pooled out-of-fold predictions show the same pattern (Figure~\ref{fig:scores}). The two models rank cases almost identically yet express that ranking differently. Pushing scores to the extremes is not an advantage: it makes the outputs look confident without making them more accurate, and it is consistent with the calibration slope below one on both partitions. Any deployment would need to ship a recalibration map with the model and to state the operating point explicitly.

\subsection{Limitations}
\label{sec:limitations}

Eleven limitations bound these conclusions. First, every experiment uses one cohort. We performed no external validation, and models of this class are known to generalise imperfectly across cohorts~\cite{fathi2024ensemble}. Second, the held-out partition holds only 20 positive pairs, so its intervals are wide; it confirms the cross-validation result rather than providing an independent estimate. Third, any recorded dementia counts as conversion, so the task is MCI-to-dementia rather than MCI-to-AD; 8 of the 133 positive pairs have a non-Alzheimer's cause. Fourth, and most limiting for the architecture, the ablations do not support the three-branch design. Any single branch performs as well as all three, and neither the granularity of the gate nor the baseline residual matters (Section~\ref{sec:branch_gate_ablation}); the gate coefficients are unstable across folds and do not explain individual predictions (Section~\ref{sec:interpretability_results}). The case for the Temporal Fusion Module rests on beating subtraction, not on its internal structure. Fifth, we did not compare against recent transformer or state-space longitudinal architectures~\cite{chen2024longformer,zhang2025cross}, so our claim is that longitudinal fusion helps, not that this design is the best of the longitudinal designs. Sixth, 508 of the 604 pairs are 13 months apart or less, so we could not test how sensitive these results are to longer intervals between scans (Section~\ref{sec:convergent}); that would need a cohort with a wider spread of intervals. Seventh, the encoder was frozen throughout. We did not try fine-tuning it jointly, which might shift the balance between the encoder's contribution and the temporal one. Eighth, the preprocessing involves trade-offs. Volumes are resampled to 1.6\,mm so that the whole brain fits the fixed network input, which gives up some spatial resolution. The registration to standard space is affine, so differences in brain shape between participants remain in the input. The within-subject alignment is rigid and does not correct for changes in scanner geometry between visits, which matter here because 41 of the 604 pairs change scanner model. Ninth, the acquisitions are heterogeneous (Section~\ref{sec:dataset} and Table~\ref{tab:acquisition}): scanners, sequences and acceleration differ between sites and, in 284 pairs, between the two scans of one participant. We applied no harmonisation: methods such as longitudinal ComBat~\cite{beer2020longitudinal} harmonise derived features, not image volumes, so they do not transfer directly to an end-to-end model, and image-level harmonisation remains untested here. Tenth, the models use imaging alone. We did not add age, sex, APOE genotype, MCI subtype or cognitive scores, although baseline MMSE on its own reaches an AUC of 0.787 in cross-validation (Section~\ref{sec:convergent}). What imaging adds to routine clinical variables, or gains in combination with them, is therefore not established. Eleventh, performance differs by scanner vendor. In cross-validation both longitudinal models discriminate less well on pairs acquired on GE systems (AUC 0.725 for TAFNet and 0.655 for CNN-LSTM; 124 pairs, 27 positive) than on Siemens systems (0.841 and 0.837; 300 pairs). Most GE scans in this cohort use the IR-(F)SPGR sequence and the others MPRAGE, so vendor and sequence cannot be separated here, and the GE subgroup is small. The result nevertheless argues against assuming that the reported performance holds equally across acquisition protocols.

\subsection{Future Directions}

A first priority is a comparison against current longitudinal architectures under this protocol. Beyond that, the breakdown reported here suggests that effort is better spent on the inputs than on the fusion operator: more modalities, longer observation windows, or more frequent scans. The best-motivated change is to treat the task as time-to-event prediction, and Section~\ref{sec:convergent} shows why. The score orders patients by severity and tracks the rate of later cognitive decline, but says nothing about when conversion will happen. A clinician asking \emph{how soon} gets no answer from the present formulation, and a fixed binary horizon discards timing information the diagnostic records already contain.

A further step follows from the limitations above: to combine the imaging score with non-imaging variables such as age, sex, APOE genotype and baseline cognition, and to report what imaging adds to them.

\section{Conclusion}
\label{sec:conclusion}

This paper proposed TAFNet, a temporal attention fusion network that predicts conversion from mild cognitive impairment to dementia from a pair of structural MRI scans, and evaluated it under a protocol designed so that the result would hold up. Conversion labels were taken from recorded diagnoses rather than enrolment category, censored participants were excluded, and the encoder pretraining pool was checked to share no participants with the evaluation cohort. A held-out test partition shared no participants with any cross-validation fold, and inference was participant-clustered throughout.

TAFNet reaches an AUC of 0.852 on the held-out partition, against 0.743 and 0.725 for conventional single-scan networks and 0.769 for a model that subtracts the two scans. Its advantage over each of these is significant in cross-validation and has the same direction on the held-out partition, where the intervals are wide. With the encoder and everything else held fixed, the follow-up scan adds $+0.041$ AUC on pooled cross-validation predictions ($p = 0.009$). How the two timepoints are combined matters: learned fusion does better than subtraction.

Against a CNN-LSTM baseline built on the same encoder TAFNet is comparable. The difference is $-0.007$ AUC on the held-out partition (95\% CI $[-0.027, +0.003]$) and $+0.018$ on pooled cross-validation predictions (95\% CI $[-0.003, +0.039]$), and calibration and operating points are equally close. We read this as a property of the task: roughly three-quarters of the improvement over the single-timepoint convolutional baseline comes from the pretrained encoder, and with only two timepoints different ways of combining its outputs reach a similar level. The internal structure of the fusion module does not carry the result: any single branch performs as well as all three, and the gate coefficients are unstable across folds and do not explain individual predictions.

The score carries clinical meaning: on both partitions it tracks the rate of later cognitive decline, a quantity never used in training. It does not say when conversion will happen, and it improves only modestly on a baseline cognitive test.

The default 0.5 decision threshold is wrong for this prevalence. Choosing operating points on validation data raises TAFNet's held-out sensitivity from 0.750 to 0.850 at a specificity of 0.723, and moves DenseNet3D-121 from a sensitivity of zero to 0.750, without changing any model. Sensitivity and specificity reported at a default threshold should be read as a property of that threshold.

Taken together, these results show that a pretrained encoder with learned temporal fusion predicts conversion from paired structural MRI well, and they also mark how much the second scan and the choice of fusion contribute. Further gains are more likely to come from richer inputs, such as additional modalities, clinical variables and longer observation, than from more elaborate ways of fusing two T1-weighted volumes.

\appendix
\setcounter{table}{0}
\section{Acquisition Protocols}
\label{app:acquisition}

Table~\ref{tab:acquisition} summarises the T1-weighted acquisitions in the two participant sets. Sequence family and acceleration are taken from the series description recorded by ADNI; a scan is counted as accelerated when the description names GRAPPA, SENSE or an accelerated protocol.

\begin{table}[H]
\centering
\caption{T1-weighted acquisitions by sequence family, acceleration and period, as number of scans (\%). Vendor and field strength are taken from the image headers.}
\label{tab:acquisition}
\begin{tabular}{llrr}
\toprule
 & & Longitudinal cohort & Pretraining pool \\
 & & ($n = 945$ scans) & ($n = 1{,}948$ scans) \\
\midrule
MPRAGE & non-accelerated & 413 (43.7) & 871 (44.7) \\
MPRAGE & accelerated & 318 (33.7) & 602 (30.9) \\
IR-(F)SPGR & non-accelerated & 120 (12.7) & 245 (12.6) \\
IR-(F)SPGR & accelerated & 94 (9.9) & 230 (11.8) \\
\midrule
Siemens & & 561 (59.4) & 1{,}051 (54.0) \\
GE & & 230 (24.3) & 552 (28.3) \\
Philips & & 154 (16.3) & 345 (17.7) \\
Field strength 3\,T & & 945 (100.0) & 1{,}948 (100.0) \\
\midrule
Acquired 2006--2010 & & 96 (10.2) & 271 (13.9) \\
Acquired 2011--2017 & & 849 (89.8) & 1{,}677 (86.1) \\
\midrule
Acquisition sites & & 54 & 59 \\
\bottomrule
\end{tabular}
\end{table}

\section*{Acknowledgements}
Data collection and sharing for the Alzheimer's Disease Neuroimaging Initiative (ADNI) is funded by the National Institute on Aging (National Institutes of Health Grant U19 AG024904). The grantee organization is the Northern California Institute for Research and Education. In the past, ADNI has also received funding from the National Institute of Biomedical Imaging and Bioengineering, the Canadian Institutes of Health Research, and private sector contributions through the Foundation for the National Institutes of Health (FNIH) including generous contributions from the following: AbbVie, Alzheimer's Association; Alzheimer's Drug Discovery Foundation; Araclon Biotech; BioClinica, Inc.; Biogen; Bristol-Myers Squibb Company; CereSpir, Inc.; Cogstate; Eisai Inc.; Elan Pharmaceuticals, Inc.; Eli Lilly and Company; EuroImmun; F.~Hoffmann-La Roche Ltd and its affiliated company Genentech, Inc.; Fujirebio; GE Healthcare; IXICO Ltd.; Janssen Alzheimer Immunotherapy Research \& Development, LLC.; Johnson \& Johnson Pharmaceutical Research \& Development LLC.; Lumosity; Lundbeck; Merck \& Co., Inc.; Meso Scale Diagnostics, LLC.; NeuroRx Research; Neurotrack Technologies; Novartis Pharmaceuticals Corporation; Pfizer Inc.; Piramal Imaging; Servier; Takeda Pharmaceutical Company; and Transition Therapeutics.

The authors acknowledge the facilities and scientific and technical assistance of the National Imaging Facility, a National Collaborative Research Infrastructure Strategy (NCRIS) capability, at Swinburne University of Technology.

\bibliographystyle{unsrtnat}
\bibliography{references}

\end{document}